\documentclass[review,12pt]{elsarticle}
\usepackage[table]{xcolor} 
\usepackage{times}
\usepackage{epsfig}
\usepackage{graphicx}
\usepackage{subfig}
\usepackage{amsmath}
\usepackage{amssymb}
\usepackage{bm}
\usepackage{url}
\usepackage{rotating}
\usepackage{verbatim}
\usepackage{epstopdf}

\usepackage{amsmath}
\usepackage{rotating}
\usepackage{url}
\usepackage{amsfonts}
\usepackage[nice]{nicefrac}
\usepackage{url}
\usepackage{float}
\usepackage{mathrsfs}
\usepackage{pgf}
\usepackage{setspace}
\usepackage{bm}

\newcommand{\E}{\mathbb{E}}

\newcommand{\A}{\mathbf{A}}

\newcommand{\B}{\mathbf{B}}

\newcommand{\F}{\mathbf{F}}
\newcommand{\G}{\mathbf{G}}

\newcommand{\ve}{\mathbf{v}}
\newcommand{\dvect}{\mathbf{d}}
\newcommand{\D}{\bm{\mathcal{D}}}

\newcommand{\w}{\mathbf{w}}

\newcommand{\diag}{\text{diag}}
\newcommand{\I}{\mathbf{I}}

\newcommand{\s}{\mathbf{s}}

\newcommand{\x}{\mathbf{x}}

\newcommand{\p}{\mathbf{p}}
\newcommand{\hessian}{\mathbf{H}}
\newcommand{\X}{\mathbf{X}}

\newcommand{\W}{\mathbf{W}}
\newcommand{\V}{\mathbf{V}}

\newcommand{\z}{\mathbf{z}}

\newcommand{\Gauss}{\mathcal{N}}

\xdefinecolor{shadecolor}{RGB}{127, 127, 255}

\def\ie{\emph{i.e.}~} 
\def\etal{\emph{et al.}~}

\newcommand{\squishlist}{
  \begin{list}{$\bullet$}
    {
      \setlength{\itemsep}{0pt}
      \setlength{\parsep}{0pt}
      \setlength{\topsep}{0pt}
      \setlength{\partopsep}{0pt}
      \setlength{\leftmargin}{1.5em}
      \setlength{\labelwidth}{1em}
      \setlength{\labelsep}{0.5em} } 
}

\newcommand{\squishend}{
  \end{list}}

\definecolor{tableShade}{gray}{0.9}
\allowdisplaybreaks

\let\svthefootnote\thefootnote

\usepackage{amssymb}

\usepackage[nodots]{numcompress}

\usepackage{lineno}

\journal{Image and Vision Computing Journal (IMAVIS)}

\begin{document}
  \begin{frontmatter}



\title{Improving Facial Analysis and Performance Driven Animation through Disentangling Identity and Expression}
 

    \author[1]{David Rim\footnotemark$^*$}
    \ead{daverim@gmail.com}
    
    \author[2]{Sina Honari\footnotemark$^*$}
    \ead{sina.honari@umontreal.ca} 
    
    \author[1]{Md Kamrul Hasan}
    \ead{md-kamrul.hasan@polymtl.ca} 
          
    \author[1]{Chris Pal}
    \ead{christopher.pal@polymtl.ca}
       
    \address[1]{D\'{e}partement de g\'{e}nie informatique et g\'{e}nie logiciel\\
      \'{E}cole Polytechnique Montr\'{e}al\\
      Montr\'{e}al, Qu\'{e}bec, Canada, H3T 1J4\\
      }
      
     \address[2]{D\'{e}partement d'informatique et de recherche op\'{e}rationnelle\\
      Universit\'{e} de Montr\'{e}al\\
      Montr\'{e}al, Qu\'{e}bec, Canada, H3C 3J7\\
      }

    \begin{abstract}      
      We present techniques for improving performance driven facial animation, emotion recognition, and facial key-point or landmark prediction using learned identity invariant representations.
      Established approaches to these problems can work well if sufficient examples and labels for a particular identity are available and factors of variation are highly controlled.
      However, labeled examples of facial expressions, emotions and key-points for new individuals are difficult and costly to obtain. 
      In this paper we improve the ability of techniques to generalize to new and unseen individuals by explicitly modeling previously seen variations related to identity and expression.  
       We use a weakly-supervised approach in which identity labels are used to learn the different factors of variation linked to identity separately from factors related to expression. 
       We show how probabilistic modeling of these sources of variation allows one to learn identity-invariant representations for expressions which can then be used to identity-normalize various procedures for facial expression analysis and animation control.  We also show how to extend the widely used techniques of active appearance models and constrained local models through replacing the underlying point distribution models which are typically constructed using principal component analysis with identity-expression factorized representations. 
       We present a wide variety of experiments in which we consistently improve performance on emotion recognition, markerless performance-driven facial animation and facial key-point tracking.
    \end{abstract}


  \end{frontmatter}

\let\thefootnote\relax\footnote{$^*$Joint first authors.}
\addtocounter{footnote}{-5}\let\thefootnote\svthefootnote






  \section{Introduction}
  
  One of the primary sources of variation in facial images is identity. Although this is an obvious statement, many approaches to vision tasks other than facial recognition do not directly account for the interaction between identity-related variation and other sources. However, many facial image datasets are subdivided by subject identity and this provides additional information that is often unused. 
  This paper deals with the natural question of how to effectively use identity information in order to improve tasks other than identity recognition. In particular, our primary motivations, applications of interest and goals are to develop methods for facial expression analysis and performance driven facial animation that are less identity dependant.

  Recently there has been work on facial recognition in which identity is separated from other sources of variation in 2D image data in a fully probabilistic way \cite{prince_probabilistic_2011}.
  In this model, the factors are assumed to be additive and independent. 
  This procedure can be interpreted as a probabilistic version of Canonical Correlation Analysis (CCA) presented by Bach \cite{bach_probabilistic_2005}, or as a standard factor analysis with a
  particular structure in the factors.
%
  In this paper, we investigate and extend the use of this probabilistic approach to separate sources of variation, but unlike prior work which focuses on inferences about identity, we focus on facial expression analysis and facial animation tasks. This includes performance-driven animation,
  emotion recognition, and key-point tracking. Our goal is to use learned representations so as to create automated techniques for expression analysis and animation that better generalize across identities. We show here how disentangling factors of variation related to identity can indeed yield improved results.
  In many cases we show how to use the learned representations as input to discriminative classification methods.
  
  In our experimental work, we apply this learning technique to a wide variety of different input types including: raw pixels, key-points and the pixels of warped images. We evaluate our approach by predicting: standard emotion labels, facial action units, and `bone' positions or animation sliders which are widely used in computer animation. We go on to show how it is indeed possible to improve the facial keypoint prediction performance of active appearance models (AAMs) on unseen identities as well as increase the performance of constrained local models (CLMs) through our identity-expression factorization extensions to these widely used techniques.  Our evaluation tasks and a comparison of the types of data being used as input for each experiment are summarized in Table \ref{tab:exp_summary}. 
  
The rest of this manuscript is structured as follows:
In Section \ref{sec:relwork}, we discuss the facial expression analysis, performance driven animation and keypoint tracking applications that serve as the ultimate goals of our work in more detail and review relevant previous work. The various methods we present here build in particular on the work of Prince et al. \cite{prince_probabilistic_2011} in which a linear Gaussian probabilistic model was proposed to explicitly separate factors of variation due to identity versus expression. 
 In section \ref{sec:basic_model}, we present this model as a way to disentangle factors of variation arising from identity and expression variation. While Prince et al. used this type of identity-expression analysis to make inferences about identity, our work here focuses on how disentangling such factors can be used to make inferences about facial expression. Indeed, as discussed above, facial expression analysis and the applications of detailed facial expression analysis to computer animation is the motivating goal of our work here. 
  
  The first set of contributions of our work are presented in section \ref{sec:FE_Expts}.  These contributions consist of a wide variety of novel techniques for using learned identity-expression representations for common goals related to expression analysis and computer animation. 
  We provide novel techniques and experiments in which we predict: emotion labels, facial action units, facial keypoints, and animation control points as summarized in Table \ref{tab:exp_summary}. 
  Of particular note is the fact that we propose a novel formulation for identity normalizing facial images which we use for facial action unit recognition, emotion recognition and animation control. We find that using this identity-normalized representation leads to improved results across this wide variety of the expression analysis tasks. 
  
We also go on to extend the underlying identity and expression analysis technique in two important directions, providing two additional technical contributions. First, in Section \ref{sec:IE-AMMs} we show that identity-expression analysis can be used in place of the principal component analysis (PCA) technique that is widely used in active appearance models (AAMs). This procedure yields what we call an identity-expression factorized AAM, or IE-AAM. We present the modifications that are necessary to integrate this approach into the AAM framework of \cite{matthews_active_2004}. Our experiments show that IE-AAMs can increase the performance of PCA-AAMs dramatically when no training data is available for a given subject. We also found that the use of IE-AAMs eliminated the convergence errors observed with PCA-AAMs. Secondly, in Section \ref{sec:IE-CLMs} we show that constrained local models (CLMs) can also be reformulated, extended and improved through using an underlying identity-expression analysis model. Our reformulation also provides a novel energy function and minimization formulation for CLMs in general.

\section{Our Applications of Interest and Relevant Prior Work}
\label{sec:relwork}

  \paragraph{Performance-driven facial animation}
  
  Performance-driven animation is the task of controlling a facial animation via images of a performer. This is a common process in the entertainment industry. 
  In many cases, this problem is generally handled by the placement of special markers on the performer~\cite{williams_performance-driven_1990}. 
  However, here we are interested in developing markerless motion capture techniques in which we employ machine learning methods and
  minimize manual intervention.

  Many marker-less facial expression analysis methods rely either on tracking points using optical flow~\cite{essa_modeling_1996} or
  fitting Active Appearance Models~\cite{lanitis_automatic_1997}. Morphable models in~\cite{blanz_morphable_1999}, \cite{pighin_resynthesizing_1999}, \cite{5457498} have also been investigated.
  More recently, 3D data and reconstruction is used to fit directly to the performer ~\cite{wang_high_2004}, ~\cite{deng_spacetime_2007}. These methods, however, often require additional data, 
  for example, multiview stereo \cite{Yin2008}, \cite{Beeler:2010:HSC:1778765.1778777} 
  or structured light \cite{Chang2005}, \cite{wang_high_2004}. Sandbach \etal provide a thorough review of 3D expression recognition techniques in \cite{Sandbach2012}. 
  In the end, these methods usually work by providing dense correspondences which then require a re-targeting step.

  However, a simpler and often used approach in industry does not rely on markers and simply uses the input video of a facial performance. The idea is to use a direct 2D to 3D mapping based on regressing image features~\cite{goudeaux_principal_2001} to 3D model parameters.
  This method works well but is insufficiently automatic. Each video is mapped to 3D model parameters, possibly with interpolation between frames. Key-point based representations 
  often require both data and training time that compares unfavorably to this simpler approach given the re-targeting step. The primary benefit of key-point based representations appears to be a degree of natural identity-invariance. We provide experiments on performance driven animation, in which we predict bone positions using the well known Japanese Female Facial Expression (JAFFE) Database \cite{lyons_coding_1998} as input as well as an experiment using professional helmet camera based video used for high quality, real world animations.

  \paragraph{Emotion recognition}

  There is a large amount of overlap in the objectives for performance-driven facial animation and the objectives of detailed emotion recognition. 
  Automatic emotion recognition has largely focused on the facial action coding system or FACS ~\cite{ekman_argument_1992} as an auxiliary task~\cite{valstar_first_2011}. In our work here we shall predict action unit (AU) values that have been annotated for the well known extended Cohn Kanade database \cite{Lucey2010}. In some cases, a FACs representation may be a useful intermediate representation for animation control.
  
 Emotions can also be encoded via the simpler and more intuitive labels for well known emotion types of   ``Anger,'' ``Disgust,'' ``Fear,'' ``Happiness,'' ``Sadness,'' ``Surprise''. We evaluate the performance of recognizing the annotations of these emotions for JAFFE  as well as CK+  in which ``Contempt'' is also included. These labels correspond to Paul Ekman's famous set of six basic emotions plus ``Contempt'' which is a part of his extended list of emotions.
 
  Many existing approaches for emotion recognition also require significant preprocessing for contrast normalization and alignment, which helps to alleviate pose variation~\cite{li_facial_2011}. Indeed, an intermediary task that has often been used as a pre-processing phase to produce emotion recognition systems is key point detection and tracking ~\cite{suwa_preliminary_1978},~\cite{bartlett_measuring_1999},~\cite{valstar_facial_2005},~\cite{cohn_spontaneous_2010}.  Tracked key-points themselves can also be used to control animations directly or indirectly via warping operations on images as we explore in our experiments on professional facial performance video obtained via helmet cameras.

  \paragraph{Key-point localization}

  Active Appearance Models (AAMs) \cite{Edwards1998}, \cite{Cootes1995} and
  Constrained Local Models (CLMs) \cite{Saragih2010} are widely used techinques for keypoint tracking. Both of these methods rely on so called point distribution models (PDMs) \cite{Cootes1995} which are constructed using principal component analysis (PCA).
  
  AAMs and CLMs usually suffer from a degree of identity-dependence. That is, a model trained on a sample of subjects does not necessarily perform well on an unseen subject. 
    AAMs in particular suffer greatly from this effect, performing much better when samples of an individual are used for both training and testing. In this paper, we investigate the problem when no additional information about a new subject is available, as is the case in many common scenarios. As such, we evaluate our method using the AAM as a test algorithm, because it performs poorly in this setting but performs well otherwise.
 
  View-based approaches \cite{Pentland1994}, and multi-stage solutions \cite{Tistarelli2009}, \cite{Liao2004} address this issue by using multiple subspaces for each identity.  Gaussian mixture models, \cite{Frey1999}
  indirectly deal with identity variation by learning clusters of training data. Gross \etal described reduced fitting robustness of AAMs on unseen subjects \cite{Gross2005}, suggesting
  simultaneous appearance and shape fitting improve generalization.
  Our method approaches identity variation directly and probabilistically, learning both factors of variation simultaneously. 

Some other notable recent work from Jeni et al. has extended CLMs to account for 3D shape. Their work has demonstrated extremely high performance on the CK+ expression classification and action unit detection evaluations. We compare our approach with their method in Section \ref{sec:FE_Expts}.

Other work has used a structured max-margin approach to keypoint placement \cite{zhu2012face} yielding state of the art results for facial keypoint placement. We compare with this approach in Section \ref{sec:IE-CLMs}.

Deep learning has recently emerged as a method capable of yielding state of the art results for keypoint placement \cite{honari2016CVPR, sun2013deep}. Deep networks are however known to overfit when data set sizes are small, which is the case for many of the problems that we are addressing in our work here (ex. markerless facial performance capture imagery from helmet mounted cameras). 

\paragraph{Other factorization techniques and our approach}
  Multi-linear and bilinear analysis of facial images \cite{Freeman1997}, \cite{heyden_multilinear_2002},~\cite{vlasic_face_2005} can model the interaction of different kinds of variation. However, with these models, a full image tensor is often needed, which can be difficult to obtain. 
  Some approaches overcome this restriction by treating the required tensor labels as missing \cite{DelBue2012}. However, this leads to discarding data. 

  The problem we address in this paper is that of learning an expression representation which leverages identity information, then using the technique to improve a wide variety of expression and emotion related tasks.
  This is similar in spirit to the expression synthesis approaches used by Du and Lin \cite{Du2003} and by Zhou and Lin \cite{Zhou2005}, which generate identity-independent expression factors. Our approach generalizes this work as a framework for unsupervised learning of expression factors, building in particular on the identity-expression factorization technique of Prince et al. \cite{prince_probabilistic_2011}. In the next section we present their probabilistic approach to this task, then go on to show how show how this formulation can be used to create an extremely wide variety of new techniques leading to improved performance for expression recognition and facial action recognition tasks. Subsequently, in Sections \ref{sec:IE-AMMs} and \ref{sec:IE-CLMs} we go on to extend the widely used formulations of AAMs and CLMs so as to use identity-expression factorized representations. In both cases we show that keypoint localization performance is improved over the traditional techniques which use principal component analysis as their underlying point distribution models.
  %

  %
  %
  \section{A Model for Disentangling Identity and Expression}
  \label{sec:basic_model}

 Face image datasets have a rich divesity due to many contributing factors of variation such as age, illumination, identity, pose, facial expression and emotion. In this literature,  we consider two main sources of variation. The primary variation is due to the identity factors and the second source of variation is due to the facial expressions as well as some degree of pose deviation.
 
\begin{figure}[htb]
    \centering
    \includegraphics[width=.4\textwidth]{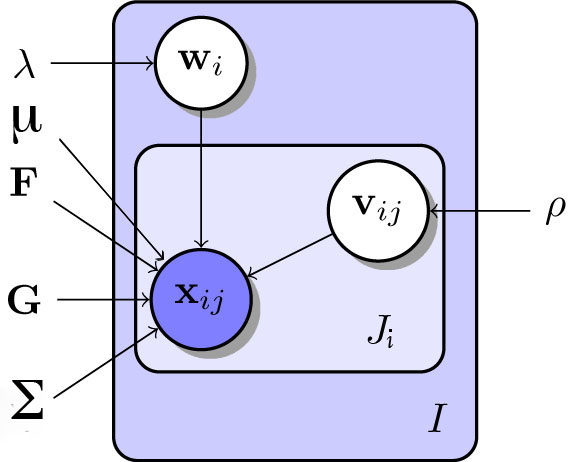}
    \caption{Graphical model of facial data generation. $\x_{ij}$ is generated from $p(\x_{ij} \: | \: \ve_{ij}, \w_i)$, after sampling
      $\w_i$ from an identity and $\ve_{ij}$ from an expression-related distribution respectively.}
    \label{fig:genmodel}
  \end{figure}

  A graphical model of this approach is shown in Figure \ref{fig:genmodel}, where each face image $\x_{ij} \in \X$ is generated by $w_i$, representing the identity $i$, and $v_{ij}$, representing the $j^\text{th}$ expression of identity $i$. The set of observed data is $\X=\{\x_{ij}\}_{i=1,...,I ~;~ j=1,...,J_i}$ which contains all images of $I$ identities, where for each each identity $i$, a total of $J_i$ expressions exist. The observed data can be pixels, vectors containing key-point locations or a concatenation of both. 
  We use the notation $\x_{ij}$ to denote the $j^\text{th}$ image of the $i^\text{th}$ identity in the dataset.

  Each $\x_{ij}$ is generated by sampling $\w_i$ and $\ve_{ij}$ from Gaussian distributions corresponding
  to identity $i$ and its $j^\text{th}$ expression distributions $p(\w_i)$ and $p(\ve_{ij})$, and then
  combining these by sampling an image $\x_{ij}$ according to $p(\x_{ij} \: |  \: \w_i, \ve_{ij})$.
  We use zero-mean independent Gaussian distributions for $p(\w_i)$ and $p(\ve_{ij})$.
  \begin{align}
    p(\w_i) &= \Gauss(\w_{i}; ~ \bf0, ~ \lambda \I), 
    \label{eq:prior1}\\
    p(\ve_{ij}) &= \Gauss(\ve_{ij}; ~ \bf0, ~ \rho \I).
    \label{eq:prior2}
  \end{align}
  The observation $\x_{ij}$ is then sampled from a multivariate Gaussian conditional distribution parameterized
  by the mean $\bm{\mu}$, matrices $\F,~ \G$ and diagonal covariance $\bf\Sigma$, such that
  \begin{equation}
    \label{eq:likelihood}
    p(\x_{ij} \: | \: \w_i, \ve_{ij}) = \Gauss(\x_{ij}; ~ \bm{\mu} + \F\w_i + \G\ve_{ij}, ~ \bf\Sigma). 
  \end{equation}
  This corresponds to a conditional distribution given the variable $\w_i$, which is identical for all images of a unique identity, and the variable $\ve_{ij}$, which varies across different expressions
  of a particular identity. The loading matrices $\F$ and $\G$ correspond respectively to identity and expression and are shared across all observations.
  The joint probability can be written as
  \begin{align}
    p(\X, \W, \V \: | \: & \F, \G, \lambda, \rho, \bm{\mu}) =
    \prod_i^I \Gauss(\w_i; ~ \bf0, ~ \lambda \I) \\
    &\prod_j^{J_i}  \Gauss(\x_{ij}; ~ \bm{\mu} + \F\w_i + \G\ve_{ij}, ~ \bf\Sigma) 
    \Gauss(\ve_{ij}; ~ \bf0, ~ \rho \I ).   \notag
  \end{align}
  where $\bf \X$ is the entire observable dataset and $\W = \{ w_1, w_2, \dots, w_I\}$ is the set of all latent identity representations and $\V = \{ v_ij\}_{i=1,...,I ~;~ j=1,...,J_i}$ is the set of all latent expression representations for all identities.
  \subsection{Learning}
  \label{sec:learning}
  The Expectation Maximization (EM) approach is used to learn the parameters of the model $\bm{\theta} = \{\F, \G, \bm{\Sigma}, \bm{\mu}, \lambda, \rho \}$. The goal is to maximize the joint distribution by alternatively maximizing and taking expectations
  \begin{equation}
    \max_{\bm{\theta}} \E[ \log  p(\X, \W, \V \: | \: \bm{\theta}) ],
  \end{equation}
  where the expectation is taken with respect to the posterior 
  distribution 
  \begin{equation}
  p(\W, \V ~ | ~ \X, \bm{\theta}^\text{old})
  \end{equation}
  and then the parameters in $\bm{\theta}$ are updated by maximizing the log of the joint distribution $p(\W, \V, \X)$.
  Since the priors in Eqs. (\ref{eq:prior1}) and (\ref{eq:prior2}) and the likelihood distribution in Eq. (\ref{eq:likelihood}) are all Gaussian, the resulting joint distribution is also Gaussian.

  For a given identity $i$, the set of observed variables $\{\x_{ij}\}_{j=1,...,J_i}$, can be written as a single feature vector 
  %
  $\x_i = (\x_{i1}^T,\x_{i2}^T,\ldots, \x_{iJ_i}^T)^T$.
  %
  Similarly the factors can be combined into a single loading matrix
  \begin{equation}
  \label{A}
    \A_i = \left( \begin{array}{ccccc} 
      \F & \G &  0 & ... &  0 \\
      \F & 0 &  \G & ... &  0  \\
      \ldots & \ldots & \ldots & \ldots & \ldots \\
      \F & 0 &  0 & ... &  \G  \\    
    \end{array} \right),
  \end{equation}
  where the number of rows and columns in $\A_i$ are $J_i$ and $J_i + 1$. The latent space representation corresponding to identity $i$ becomes
  \begin{equation}
    \dvect_i = (\w_{i}^T, \ve_{i1}^T,\ve_{i2}^T,\ldots,\ve_{iJ_i}^T)^T.
  \end{equation}

  Having $\w_i \in R^K$ and $\ve_{ij} \in R^{L}$, $\dvect_i$ becomes a vector of dimensionality $K + L \times J_i$. Since $\dvect_i$ is composed of two sets of vectors
  with zero-mean Gaussian distributions, it is also distributed as a zero mean Gaussian: 
  \begin{align}
    p(\dvect_i) & = \Gauss(\dvect_i; \: \bf0, \: {\bf \Phi}_i), \\
    {\bf \Phi}_{i} & = diag\left(\lambda_{1},~\ldots,~\lambda_{K},~\rho_{1{}_{1}},~\ldots,~\rho_{1{}_{L}},~\ldots,~\rho_{Ji_{1}},~\ldots,~\rho_{Ji_{L}}\right), 
  \end{align}
where the term ${\bf \Phi}_i$ is a diagonal covariance matrix, in which the first $K$ elements of the diagonal are extracted from the diagonal elements of the covariance matrix $\lambda \I$. The diagonal of ${\bf \Phi}_i$ is then composed of $J_i$ blocks of $L$ elements that are extracted from the diagonal of covariance matrix $\rho \I$ repeated $J_i$ times. Intuitively the first set of $K$ elements represent identity variations, while each of the $J_i$ blocks represent $L$ factors encoding expression variations within each image. Given this construction, the probability for identity $i$ can be rewritten as a Gaussian
  $p(\x_i ~ | ~ \dvect_i) =  \Gauss(\x_i; ~ \bm{m}_i + \A_i\dvect_i, ~ \bm{\Psi}_i)$, where $\bm{\Psi}_i$ is constructed as a diagonal matrix by concatenating $J_i$ times the diagonal of $\Sigma$. The term $\bm{m}_i$ is $J_i$ blocks of $\bm{\mu}$ being concatenated.  
   
   The posterior probability of $\dvect_{i}$ is also Gaussian with moments
  \begin{align}
    \label{eq:posterior1}
    \E[\dvect_i] &= ({\bf \Phi}_i^{-1} + \A_i^T{\Psi_i}^{-1}\A_i)^{-1}\A_i^T{\Psi_i}^{-1}(\x_i - \bm{m}_i), \\ 
    \label{eq:posterior2}
    \E[\dvect_i\dvect_i^T] &= ({\bf \Phi}_i^{-1} + \A_i^T{\Psi_i}^{-1}\A_i)^{-1} - \E[\dvect_i]\E[\dvect_i^T].
  \end{align}

  In the E-step, the mean and covariance matrices of the posterior distribution $p(\dvect_i ~ | ~ \x_i)$ is taken, which is measured over all of the expressions corresponding to the same identity. This representation style assures that all of the expressions of the same person have the same representation for identity. Note that the expectations in Eqs. (\ref{eq:posterior1}) and (\ref{eq:posterior2}) should be taken separately for all of the identities in the training set. In the M-step, however, we can disentangle the $\dvect_i$ into a set of $\dvect_{ij}$, in which each $\dvect_{ij}$ contains one sample of identity and one sample of expression. This is due to the fact that the parameters of the model should be updated with respect to all of the data in the training set and having done the E-step, we have already got the same identity representation for all expression observations from the same person. Therefore, disentangling $\dvect_i$ into a set of $\dvect_{ij}$ can reserve that information while at the same time it encodes a simpler latent representation for updating the parameters of the joint distribution. We define
${\dvect_{ij}}=\left[\begin{array}{c}
\w_{i}\\
\ve_{ij}
\end{array}\right]$ and ${\B}=[\begin{array}{cc}
\F & \G \end{array}]$, such that $d_{ij}$ is distributed as
  \begin{align}
    p(\dvect_{ij}) & = \Gauss(\dvect_{ij}; \bf0, {\bf \Phi}), \\
    {\bf \Phi} & = diag\left(\lambda_{1},\ldots,\lambda_{K},\rho_{1},\ldots,\rho_{L}\right).
  \end{align}

Given this notation, the conditional distribution can now be written as  
  \begin{equation}
    p(\x_{ij}|{\dvect_{ij}}) = \Gauss(\x_{ij}; {\bm{\mu}} + {\B \dvect_{ij}}, \bf\Sigma).
  \end{equation}
  
  We can further simplify this notation by setting 
   $\tilde{\dvect_{ij}}=\left[\begin{array}{c}
\dvect_{ij}\\
1
\end{array}\right]$ and $\tilde{\B}=[\begin{array}{cc}
\B & \bm{\mu} \end{array}]$, which in turn gives the following conditional and prior distributions
  \begin{equation}
    p(\x_{ij}|\tilde{\dvect_{ij}}) = \Gauss(\x_{ij}; \tilde{\B}\tilde{\dvect_{ij}}, \bm{\Sigma}), 
  \end{equation}
\begin{align}
    p(\tilde{\dvect_{ij}}) & = \Gauss(\tilde{\dvect_{ij}}; \bf0, \tilde{\bm{\Phi}})
\end{align}
with $\tilde{\bm{\Phi}}$ being equal to
\begin{align}
\tilde{\bm{\Phi}} =\left[\begin{array}{cc}
\bm{\Phi} & \bf0_{K+L, 1}\\
\bf0_{1, K+L}  &  \bf0_{1, 1}
\end{array}\right],
\end{align}
 where two zero vectors with row or column size of $K+L$ are concatenated with $\bm{\Phi}$ and another zero element to build a square matrix of size $K+L+1$. The joint distribution then gets equal to
  \begin{align}
    p(\X, \D |& \bf\Sigma, \tilde{B}, \tilde{\bf \Phi}) = 
    \prod_{ij}  \Gauss(\x_{ij}; \tilde{B}\tilde{d_{ij}}, \bf\Sigma) \Gauss(\tilde{d_{ij}}; \bf0, \tilde{\bf \Phi}). 
  \end{align}
  with $\X$ representing the whole set of observed data and $\D$ representing the entire set of latent variables. This simplification is preferred since the variables $\F, \G, \bm{\mu}$ are mutually dependent. Updating each one requires the other two. In the new notation, all of them can be updated simultaneously by updating $\tilde{\B}$. Maximizing with respect to $\tilde{\B}, {\bf \Sigma}, \tilde{\bf \Phi}$, give the following updates
  \begin{align}
    \tilde{\bf \Phi} & = \frac{1}{N}
    \sum_{ij}{\diag}\left\{\mathbb{E}\left[\tilde{\dvect}_{ij}\tilde{\dvect}_{ij}^{T}\right]\right\}, \\
    \bm{\Sigma} & = \frac{1}{N}
   \sum_{ij}{\diag}
   \left\{\tilde{\B}\mathbb{E}\left[\tilde{\dvect}_{ij}\tilde{\dvect}_{ij}^{T}\right]\tilde{\B}^{T}+\x_{ij}\x_{ij}^{T}-2\x_{ij}\mathbb{E}\left[\tilde{\dvect}_{ij}^{T}\right]\tilde{\B}^{T}\right\}, \\
   \tilde{\B} &=\sum_{ij}\left\{ \x_{ij}\mathbb{E}\left[\tilde{\dvect}_{ij}^{T}\right]\right\} \left\{ \sum_{ij}\mathbb{E}\left[\tilde{\dvect}_{ij}\tilde{\dvect}_{ij}^{T}\right]\right\}^{-1}.
  \end{align}
  Note that the parameters are updated with respect to all expressions of all identities. As for inference at test time, the procedure for determining the optimal $\dvect_{ij}$ vector is straight-forward, using the following posterior distribution:
  \begin{align}
    p(\dvect_{ij} |\x_{ij}) = \Gauss(\dvect_{ij}; ({\bf \Phi}^{-1} + \B^T{\bm{\Sigma}}^{-1}\B)^{-1}\B^T{\bm{\Sigma}}^{-1}(\x_{ij} - \bm{\mu}), ({\bf \Phi}^{-1} + \B^T{\bm{\Sigma}}^{-1}\B)^{-1}).
  \label{posterior}
  \end{align}
  %

  
  \subsection{Facial Expression Analysis, Animation Control Methods and Experiments}
  \label{sec:FE_Expts}
\begin{table}[htb]
\centering
\caption{Summary of experiments described in this paper, numbers of the section describing each experiment are given in parenthesis.}
\vspace{10pt}
\small
\begin{tabular}{lrl}
\multicolumn{3}{c}{} \\
\cline{1-3}
\multicolumn{3}{l}{JAFFE \rule{0pt}{3.6ex}} \\
\cline{1-3}
\multicolumn{3}{l}{~~ \bf{Emotion Recognition} \rule{0pt}{2.8ex}} (\ref{sec:jaffe_emotion})\\
&  \emph{predicts:} & Emotion labels   \\
& \emph{using:} & Images \\
\multicolumn{3}{l}{~~ \bf{Animation Control} \rule{0pt}{2.8ex}} (\ref{sec:jaffe_mc}) \\
& \emph{predicts:} & Bone position parameters   \\
& \emph{using:} & Images  \\
\cline{1-3}
\multicolumn{3}{l}{Extended Cohn-Kanade \rule{0pt}{3.6ex}}   \\
\cline{1-3}
\multicolumn{3}{l}{~~ \bf{Facial Action Unit (AU) Detection} \rule{0pt}{2.8ex}} (\ref{sec:ck_au}) \\
&    \emph{predicts:} & AU labels   \\
&   \emph{using:} & Point locations \\
&    & Shape-normalized images \\
&    & \begin{minipage}[t]{0.5\columnwidth} Combined point locations and shape-normalized images \end{minipage} \\
\multicolumn{3}{l}{~~ \bf{Emotion Recognition} \rule{0pt}{2.8ex}} (\ref{sec:ck_em}) \\
&    \emph{predicts:} & Emotion labels   \\
&   \emph{using:} & Point locations \\
&    & Shape-normalized images \\
&    & \begin{minipage}[t]{0.5\columnwidth} Combined point locations and shape-normalized images \end{minipage} \\
\multicolumn{3}{l}{~~ \bf{Key-Point Localization (IE-AAMs)} \rule{0pt}{2.8ex}} (\ref{sec:ck_pts}) \\
&    \emph{predicts:} & Key-point locations   \\
&   \emph{using:} & Images \\
%
\cline{1-3}
\multicolumn{3}{l}{CMU Multi-PIE \rule{0pt}{3.6ex}}    \\
\cline{1-3}
\multicolumn{3}{l}{~~ \bf{Key-Point Localization (IE-CLMs)} \rule{0pt}{2.8ex}} (\ref{sec:IE-CLMs-expts}) \\
&    \emph{predicts:} & Key-point locations   \\
&   \emph{using:} & Images \\
\cline{1-3}
\multicolumn{3}{l}{Animation Control Studio Data \rule{0pt}{3.6ex}}    \\
\cline{1-3}
\multicolumn{3}{l}{~~ \bf{Animation Control} \rule{0pt}{2.8ex}} (\ref{sec:motion_capture_studio_data}) \\
&    \emph{predicts:} & Bone position parameters   \\
&   \emph{using:} & Shape-normalized images \\
\cline{1-3}
\end{tabular}
\label{tab:exp_summary}
\end{table}
  In the following sub-sections, we build upon the model above to create techniques for emotion recognition and performance-driven animation and evaluate the methods using multiple datasets. We first present emotion 
  recognition methods and results, including the task of facial action unit classification. Then, we move on to performance-driven animation techniques and experiments. 

A summary of these experiments is shown in Table \ref{tab:exp_summary}.
  %
  %
  \subsubsection{Emotion Recognition with JAFFE}\label{sec:jaffe_emotion}

  \begin{figure}[htb]
    \centering
    \includegraphics[width=.5\textwidth]{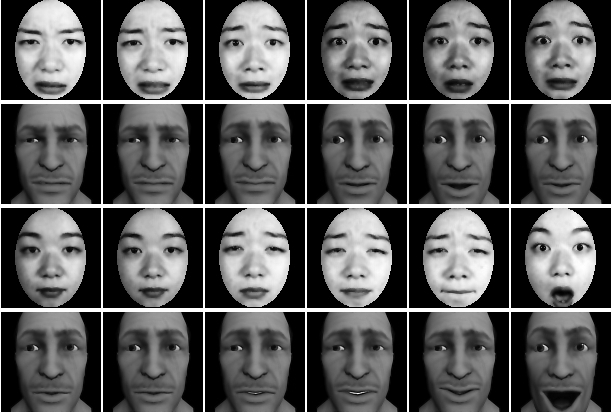}
    \caption{{The JAFFE dataset contains 213 labeled examples for 10 subjects. Images for a single test subject, left out from training, shown here with predicted labels from our method.
	The data is shown in two sets of rows. The first row in each set is the original input data, the second a rendering of the mesh with corresponding bone positions predicted by the model.}} 
    \label{fig:ex33}
  \end{figure}  
  We first run a set of experiments on the constrained JAFFE dataset~\cite{lyons_coding_1998}.
  The JAFFE dataset contains ten identities with varying expression across seven emotions containing 213 images in total. The frontal facial images are roughly aligned but we use a funneling algorithm~\cite{huang_unsupervised_2007} after
  face detection to correct small pose variation. 
  
  We then use an ellipse, manually specified, to mask background variation and reduce dimensionality, and then divide the face into three rectangular regions roughly corresponding to the mouth, eyes and ears.
  However, unlike previous works, such as \cite{Zhou2005}, we do not manually locate facial landmarks.
  Then we run the algorithm for each region individually, learning a composite space in which facial expressions are treated as a linear combinations.
  The expression space weights for each region are learned separately and then concatenated to form a single input vector consisting of the weights $\ve_{ij}$. 

  We then predict seven emotional states -- anger, disgust, fear, happiness, neutral, sadness and surprise -- using the learned representation $\ve_{ij}$.
  We train a SVM RBF classifier for each of seven emotional states on the in-sample subject images, using a single image from
  each of the emotions of the left-out subject as a validation set to learn the slack and kernel bandwidth parameters. We treat the classification at test time as a one-against-all prediction. We compare our
  results against PCA performed on the same input data as our model, using 30 and 100 dimensions. The final rates are shown in Table \ref{tab:quant} as the average accuracy across identities.

  The overall prediction accuracy rate for the emotion prediction for an unseen identity is 72.17\% using a 30 dimensional expression space. 
  A recent result on the JAFFE set by Cheng \etal~\cite{fei_cheng_facial_2010}, in experiments with left-out subjects, obtained a mean accuracy of 55.24\% using Gaussian process
  classification. Because their pre-processing is not identical and they do not report standard errors, we do not include it in Table \ref{tab:quant}.
\begin{table}
  \centering  \
  \caption{Accuracy for JAFFE emotion recognition in percentage and 
    Mean Squared Error for bone position recovery experiments for JAFFE and Studio Motion Capture data, calculated per bone position, which lie in $[-1, 1]$.}
  \vspace{5pt}
  \footnotesize
  \begin{tabular}{l c c c c c c}
    & \multicolumn{4}{c}{JAFFE} & \multicolumn{2}{c}{Studio Data} \\
    & \multicolumn{2}{c}{Emotion Recognition} & \multicolumn{2}{c}{Bone Position Recovery} & \multicolumn{2}{c}{Bone Position Recovery} \\
    & Accuracy  & SE & MSE  & SE  & MSE  & SE \\
    \hline
    No Factor Analysis & 53.13\% & 3.39\%  & 1.7526 & 0.2702 & 1.5809 & 0.2300 \\
    PCA 100 dimensions & 57.08\% & 6.57\% & 1.7526 & 0.2702 & 0.0786 & 0.0120 \\
    PCA 30 dimensions & 56.13\% & 5.64\% & 0.1223 & 0.0121  & 0.1007 & 0.0123 \\
    Our Method & \bf{72.71}\% & 1.83\% & \bf{0.0851} & 0.0077 & \bf{0.0231} & 0.0021  \\   
    \hline
  \end{tabular}
  \label{tab:quant}
\end{table}
  %
  \subsubsection{Emotion Recognition and Facial Action Unit Detection with CK+}

  Certainly, the JAFFE data exhibits far less variation than is usually present in real-world data.
  In this section, we present more detailed experiments using the Extended Cohn Kanade (CK+) database \cite{Lucey2010} for emotion recognition and facial action unit tasks.
  The CK+ dataset consists of 593 image sequences from 123 subjects
  ranging in age from 18 to 50, 69\% of whom are female and 13\% of whom are black. The images are frontal images of posed subjects taken from video sequences.
  Each sequence contains a subject posing a single facial expression starting from a neutral position. The sequence consists
  of sampled frames from videos in which the final posed position is labeled with FACS action units. In addition, emotion labels, consisting of the expressions
  ``Anger,'' ``Disgust,'' ``Fear,'' ``Happiness,'' ``Sadness,'' ``Surprise'' and ``Contempt'', are provided for 327 of the 593 sequences.

  \paragraph{Facial Action Unit Detection}\label{sec:ck_au}

  The CK+ database contains 593 labeled sequences, however only the final image as well as the initial neutral image may be used for traditional classification.
  Lucey \etal \cite{Lucey2010} describe their baseline approach based on linear classification of key-points and warped images obtained by Active Appearance Models.
  The resulting landmark data is also provided in this dataset. To compare our method with the baseline, we 
  recreate their approach. First, we use a Procrustes analysis using affine transformation of the landmark positions to determine a mean shape and register the point locations to the mean by estimating the least
  square best 2D transformation.
  We then run PCA on the difference between the Procrustes aligned points and the original points to determine the similarity components. We add these to the AAM shape parameters to model rigid motion. 
  Finally, a piece-wise affine warp is applied to normalize the shape of each facial
  image to the base shape recovered from the Procrustes analysis to obtain warped images.

  Using this approach, we generate two feature sets. The first is point locations after Procrustes analysis. The second set contains shape-normalized images, which are converted to vectors. We 
  use leave-one-subject-out cross-validation using linear SVM's for each of the 17 AU's on the vectors of landmark locations and vector of shape-normalized images, 
  and record the AUC score for each left-out subject and AU pair. An estimate for the 
  AUC error is calculated as well, defined as $\sqrt{\frac{A(1 - A)}{\text{min}(N_p, N_n)}}$, where $A$ denotes the AUC score and $N_p, N_n$ are the number of positive and negative examples
  respectively. To combine feature sets, Lucey \etal 
  run logistic regression on the scores of SVM's built from the two feature-sets independently. We also recreate this step.
  
  For our identity-normalization experiments, we use the point-locations and shape-normalized images as inputs to our factor analysis. In this case, we use 100 and 30 dimensions for the identity
  and expression parameter vectors respectively. In order to avoid overfitting, we increase the training data size for this unsupervised step from 1186 to 2588 by ensuring that
  each identity has at least 20 images, by sampling uniformly from intermediate frames. We do not use any testing subject images during training our models, including during the factor analysis in order
  to maintain fairness. 
  \begin{table}[htb]		
\rowcolors{3}{blue!15}{white}
\centering
\caption{\small{CK+: AUC Results and estimated standard errors of the AU experiment}}
\vspace{5pt}
\footnotesize
\begin{tabular}{c c c c c c c c}
\hiderowcolors 
\multicolumn{8}{c}{\bf{CK+ Facial Action Unit Recognition}} \\		
   &   & \multicolumn{3}{c}{Baseline \cite{Lucey2010}}    & \multicolumn{3}{c}{Identity Normalized} \\	
AU & N &  SPTS &  CAPP  & SPTS+CAPP  &  SPTS &  CAPP  & SPTS+CAPP \\ 		
\hline		
\showrowcolors 
1 & 173 & 94.1 $\pm$ 1.8 &  91.3 $\pm$ 2.1  & 96.9 $\pm$ 1.3  & 97.7 $\pm$ 2.8 &  96.33 $\pm$ 3.2  & \bf{99.1 $\pm$ 1.2} \\  
2 & 116 & 97.1 $\pm$ 1.5 &  95.6 $\pm$ 1.9  & 97.9 $\pm$ 1.3   & 97.2 $\pm$ 2.4 &  97.82 $\pm$ 1.0  & \bf{98.5 $\pm$ 0.9} \\  
4 & 191 & 85.9 $\pm$ 2.5 &  83.5 $\pm$ 2.7  & 91.0 $\pm$ 2.1    & 89.6 $\pm$ 9.3 &  91.72 $\pm$ 7.7  & \bf{94.3 $\pm$ 5.6} \\  
5 & 102 & 95.1 $\pm$ 2.1 &  96.6 $\pm$ 1.8  & 97.8 $\pm$ 1.5  & 96.1 $\pm$ 2.1 &  97.52 $\pm$ 2.6  & \bf{98.0 $\pm$ 1.6} \\  
6 & 122 & 91.7 $\pm$ 2.5 &  94.0 $\pm$ 2.2  & 95.8 $\pm$ 1.8   & 96.8 $\pm$ 3.3 &  95.37 $\pm$ 4.1  & \bf{97.3 $\pm$ 2.8} \\  
7 & 119 & 78.4 $\pm$ 3.8 &  85.8 $\pm$ 3.2  & 89.2 $\pm$ 2.9   & 89.7 $\pm$ 7.4 &  92.95 $\pm$ 7.1  & \bf{94.7 $\pm$ 5.9} \\  
9 & 74 & 97.7 $\pm$ 1.7 &  99.3 $\pm$ 1.0  & 99.6 $\pm$ 0.7   & 99.2 $\pm$ 0.8 & \bf{99.71 $\pm$ 0.4} & 98.2 $\pm$ 0.2 \\  
11 & 33 & 72.5 $\pm$ 7.8 &  82.0 $\pm$ 6.7  & 85.2 $\pm$ 6.2   & 81.1 $\pm$ 4.1 &  83.15 $\pm$ 4.2  & \bf{91.6 $\pm$ 3.6} \\  
12 & 111 & 91.0 $\pm$ 2.7 &  96.0 $\pm$ 1.9  & 96.3 $\pm$ 1.8    & 96.9 $\pm$ 2.2 & \bf{97.68 $\pm$ 2.2} & 97.0 $\pm$ 2.0 \\  
15 & 89 & 79.6 $\pm$ 4.3 &  88.3 $\pm$ 3.4  & 89.9 $\pm$ 3.2   & 90.2 $\pm$ 4.2 &  96.42 $\pm$ 2.0  & \bf{96.8 $\pm$ 2.5} \\  
17 & 196 & 84.4 $\pm$ 2.6 &  90.4 $\pm$ 2.1  & 93.3 $\pm$ 1.8   & 92.1 $\pm$ 6.7 &  95.48 $\pm$ 4.2  & \bf{96.3 $\pm$ 3.6} \\  
20 & 77 & 91.0 $\pm$ 3.3 &  93.0 $\pm$ 2.9  & 94.7 $\pm$ 2.6  & 95.9 $\pm$ 3.4 &  94.62 $\pm$ 2.4  & \bf{96.7 $\pm$ 1.7} \\  
23 & 59 & 91.1 $\pm$ 3.7 &  87.6 $\pm$ 4.3  & 92.2 $\pm$ 3.5   & 91.7 $\pm$ 3.7 &  93.77 $\pm$ 2.9  & \bf{96.5 $\pm$ 2.8} \\  
24 & 57 & 83.3 $\pm$ 4.9 &  90.4 $\pm$ 3.9  & 91.3 $\pm$ 3.7   & 88.9 $\pm$ 4.5 &  92.91 $\pm$ 3.4  & \bf{94.2 $\pm$ 3.4} \\  
25 & 287 & 97.1 $\pm$ 1.0 &  94.0 $\pm$ 1.4  & 97.5 $\pm$ 0.9   & 97.6 $\pm$ 2.6 &  96.83 $\pm$ 3.6  & \bf{98.1 $\pm$ 1.9} \\  
26 & 48 & 75.0 $\pm$ 6.3 &  77.6 $\pm$ 6.0  & 80.3 $\pm$ 5.7  & 77.9 $\pm$ 8.1 &  83.61 $\pm$ 7.4  & \bf{86.6 $\pm$ 6.8} \\  
27 & 81 & 99.7 $\pm$ 0.7 &  98.6 $\pm$ 1.3  & 99.8 $\pm$ 0.5    & 99.9 $\pm$ 0.2 &  99.67 $\pm$ 0.5  & \bf{99.9 $\pm$ 0.2} \\  
\hline   
\hiderowcolors 
AVG &   & 90.0 $\pm$ 2.5 &  91.4 $\pm$ 2.4  & 94.5 $\pm$ 2.0        & 92.8 $\pm$ 4.0 &  94.45 $\pm$ 3.5  & \bf{96.1 $\pm$ 2.7} \\   
\hline  
\end{tabular}  
\label{tab:au_results}  
\end{table}  

  The results are shown in Table \ref{tab:au_results}, which shows an increase in average AUC for each of the feature-sets individually. In a majority of AU classes, the AUC score increased, while
  there are no large reductions in AUC after identity normalization. For both the SPTS and CAPP scores, identity normalization showed significant increases in performance for AU-7, Lid-Tightener
  and AU-15, Lip Corner Depresser. AU-7 is associated with an eye-narrowing or squinting action, and AU-15. We show examples from this dataset of these two AU's to show that resulting
  normalization leads to representations that can be seen as roughly identical.

  In both cases, the expression is well reproduced in the identity-normalization procedure.
  The increase in performance from using both feature-sets is apparent again, which is greater than .86 AUC for every AU and well over 90\% for most of them.
  Our comparison shows that the simple step of identity normalization improves performance on average. It appears to work especially well for those AU's that are 
  easy, in some sense, to capture between identities.

 We also place our work in further context with the method recently proposed by Jeni et al. \cite{Jeni} for shape based action unit recognition. They give an average AUC result of 89.13\%. Our shape, texture and combined average AU results were 92.8\%, 94.4\% and 96.1\%, respectively.
 
  \paragraph{Emotion Detection}\label{sec:ck_em}
  
  The baseline approach can also be used for emotion recognition. Again, we recreate their approach and use the identity-normalization step as an unsupervised learning procedure.
  Because of this, we can use more training examples than those that are labeled, making this procedure semi-supervised. Again, we use the 2588 images used in the AU experiments.
  Each of the 327 labeled examples along with the neutral examples are then projected into the space learned in the previous step and used as input for training the linear SVM's. 
  In this case, each SVM learns a one-vs-all binary classifier for the emotion of interest, using the rest as negative examples. The multi-class decision is made using the maximum score. In order
  to recreate the experiments in \cite{Lucey2010}, the neutral examples are left out of the testing, resulting a forced-choice between the seven emotions. Again, the SVM's and logistic regressors are 
  learned using leave-one-subject-out, that is, over a total of 123 trials.

 \begin{table}[hbt]
\rowcolors{2}{blue!15}{white}
\centering
\footnotesize
\caption{\small{
Comparison of confusion matrices of emotion detection for the combined landmark (SPTS) and shape-normalised 
image (CAPP) features before and after identity normalisation. The average accuracy for all predicted
emotions using the state of the art method in \cite{Lucey2010} is 83.27\% (top table), using our method yields 95.21\% (bottom table), a substantial improvement of 11.9\%.
}}
\vspace{5pt}
\begin{tabular}{c c c c c c c c}
\hiderowcolors 
\multicolumn{8}{c}{\bf{CK+ Emotion Recognition}} \\
\multicolumn{8}{c}{Baseline SPTS+CAPP \cite{Lucey2010}} \\
  &  An &  Di &  Fe &  Ha &  Sa &  Su &  Co \\
\cline{2-8}
\showrowcolors
\multicolumn{1}{r|}{An}&  \bf{75.0} &  7.5  &  5.0  &  0.0  &  5.0  &  2.5  &  5.0 \\
\multicolumn{1}{r|}{Di}&  5.3  & \bf{94.7} &  0.0  &  0.0  &  0.0  &  0.0  &  0.0 \\
\multicolumn{1}{r|}{Fe}&  4.4  &  0.0  &  \bf{65.2} &  8.7  &  0.0  &  13.0  &  8.7 \\
\multicolumn{1}{r|}{Ha}&  0.0  &  0.0  &  0.0  & \bf{100.0} &  0.0  &  0.0  &  0.0 \\
\multicolumn{1}{r|}{Sa}&  12.0  &  4.0  &  4.0  &  0.0  &  \bf{68.0} &  4.0  &  8.0 \\
\multicolumn{1}{r|}{Su}&  0.0  &  0.0  &  0.0  &  0.0  &  4.0  & \bf{96.0} &  0.0 \\
\multicolumn{1}{r|}{Co}&  3.1  &  3.1  &  0.0  &  6.3  &  3.1  &  0.0  &  \bf{84.4} \\
\hiderowcolors
\multicolumn{8}{c}{\vspace{5pt}} \\
\multicolumn{8}{c}{Identity Normalized SPTS+CAPP} \\
&  An &  Di &  Fe &  Ha &  Sa &  Su &  Co \\
\showrowcolors
\cline{2-8}
\multicolumn{1}{r|}{An}&  \bf{95.4} &  0.0  &  0.0  &  0.0  &  0.0  &  0.0  &  4.7 \\
\multicolumn{1}{r|}{Di}&  1.9  &  \bf{94.2} &  0.0  &  0.0  &  1.9  &  0.0  &  1.9 \\
\multicolumn{1}{r|}{Fe}&  0.0  &  0.0  &  \bf{84.2} &  0.0  &  0.0  &  5.3  &  10.5 \\
\multicolumn{1}{r|}{Ha}&  0.0  &  0.0  &  1.6  &  \bf{96.9} &  0.0  &  0.0  &  1.6 \\
\multicolumn{1}{r|}{Sa}&  0.0  &  0.0  &  5.0  &  0.0  &  \bf{90.0} &  0.0  &  5.0 \\
\multicolumn{1}{r|}{Su}&  0.0  &  0.0  &  0.0  &  0.0  &  0.0  &  \bf{98.8} &  1.2 \\
\multicolumn{1}{r|}{Co}&  7.7  &  0.0  &  0.0  &  0.0  &  0.0  &  0.0  &  \bf{92.3} \\
\end{tabular}
\label{tab:confusions_both2}
\end{table}  
  
  However, after combining scores from the SVM's using a logistic regression technique, the improvements are quite impressive. As in \cite{Lucey2010}, results after the calibration of SVM scores indicate that the point data and
  image data capture different kinds of information. Combining these features yields impressive gains in performance. ``Anger,'' especially improves using our method. In particular, only ``Happiness'' recognition
  is reduced using our approach, but the reduction in hit rate is modest. Overall, this method gave a 95.21\% average accuracy compared to the result of 83.27\% reported in \cite{Lucey2010}.
  Comparison of confusion matrices of emotion detection for the combined landmark (SPTS) and shape-normalised 
  image (CAPP) features before and after identity normalisation is shown in Table \ref{tab:confusions_both2}.
  
More recently Jeni et al. \cite{Jeni} have proposed a 3D shape model and explored two normalization methods for the CK+ expression recognition evaluation. They report that their action unit 0 or AU0 normalization procedure yields an average class accuracy of 96\%, while their personal mean shape procedure yielded an average class accuracy of 94.8\%. Our average accuracy above is given as the per example average whereas these results were computed on a per-class basis. Our average per-class accuracy was 93.1\%.

  
  \subsubsection{Animation Control Experiments with JAFFE}\label{sec:jaffe_mc}

  We again begin using the JAFFE dataset. Each image, after face detection and alignment, is labeled using a common 3D face mesh fitted with 27 bones created by an artist, as can be seen in Figure \ref{fig:ex33}.
  Each bone is then positioned by an artist to a maximal and minimal position along a fixed path. The position along the path is specified by a real value in $[-1, 1]$ as a fraction of the distance between the 
  midpoint and the extreme values.

  We use MSE for evaluation purposes on the animation experiment, which corresponds to bone parameter recovery. In this case, we use linear regression as the predictive algorithm.
  The MSE reported is the average error in bone position for all test images.
  For each trial, we leave one identity out from the trial, and compute both the MSE error and a prediction for the facial action. We compare our approach
  to PCA, using 30 and 100 dimensions. The results are summarized in Table \ref{tab:quant}. For the experiment labeled ``none,'' the experiment denotes no unsupervised pre-processing step -- the input is the raw image data. 
  As it is difficult to gauge the quality of the predictions from MSE alone, the predictions
  for a test subject using our method are illustrated in Figure \ref{fig:ex33}, showing that the method does recover the facial actions produced by the subject quite well. The small standard error of the MSE indicates
  that, in general, the procedure is capable of predicting facial actions across all unseen subjects. 
  %
  %
  \subsubsection{Animation Control Experiments with Markerless Motion Capture}\label{sec:motion_capture_studio_data}
\begin{figure}[htb]
    \centering
    \subfloat[Original data]{
      \includegraphics[height=15mm]{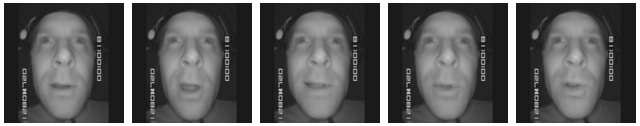}
    }\\
    \hspace{-0.5em}
    \centering
    \subfloat[Results of AAM model]{
      \includegraphics[height=15mm]{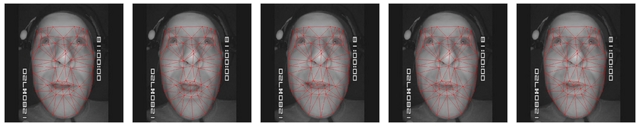}
      \label{fig:expb}
    }\\
    \hspace{-0.5em}
    \centering
    \subfloat[Piecewise warped data]{
      \includegraphics[height=15mm]{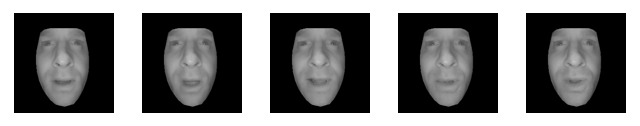}
    }\\
    \hspace{-0.5em}
    \centering
    \subfloat[Identity normalization is applied]{
      \includegraphics[height=15mm]{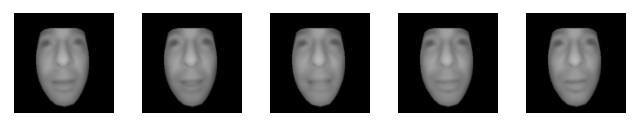}
    }
    \caption{Motion capture training data using a helmet IR camera. Example pipeline for a test video sequence.}
    \label{fig:exp}
  \end{figure}

  We now return to the challenging real-world problem of high quality facial animation control using video obtained from helmet cameras. This technique was used in the well known film Avatar and this data comes from Ubisoft, the company responsible for the brands Assassin's Creed and FarCry among others. FarCry 3 uses these technologies extensively.  We obtained 16 videos of motion capture data without marker data. These videos are produced using infra-red cameras.
  Examples are shown in Figure \ref{fig:exp}. As is common in motion capture data, faces are captured using a helmet-mounted camera which reduces pose variation. The camera position is relatively fixed.
  However, this data presents new challenges due to variation stemming from the varied appearance of the actors. To compensate for appearance variation due to facial structure, an Active Appearance
  Model is applied to the data as shown in Figure \ref{fig:exp}, using 66 fiducial points. The resulting points derived from the model are used to warp each video frame to a common coordinate structure, 
  removing pixels from outside the convex hull. These were used as inputs to our model, for a dataset totaling 3122 frames. Example outputs can also be seen in Figure \ref{fig:expb}.

  For our evaluations, we were given professional results of a bone-based model used as motion capture output. The bone positions are given values in $[-1, 1]$ representing the relative distance
  from the neutral position for each bone along predefined space. Each experiment is run with one subject left out, and the results shown are the average MSE and standard error. Results of the 
  experiments are shown in the right-most column of Table \ref{tab:quant}. We first show results using no preprocessing except for face detection
  using the Viola-Jones detector, cropped to a size 1.5 time the size of the detector to account for the distortion evident in the images. The results shown in Table \ref{tab:quant} are per bone, and 
  clearly some data processing must be used (None* refers to the images after shape normalization). Best results were not obtained by applying PCA to the original data. The remaining experiments
  used shape-normalised data as inputs. As can be seen PCA improves the results, but the best results are achieved using our model. The classifier used was a simple linear regression, 
  a more powerful discriminative classifier may be more effective.
  %
  %
  \section{Identity-Expression Active Appearance Models (IE-AAMs)}
  \label{sec:IE-AMMs}
  For many of the experiments in the preceding section a key step is in the application of key-point detection algorithms, specifically AAMs.
  The AAM has been shown to work very well for subjects on which it has been trained while
  generalization to unseen identities has proven more difficult, due to the increased complexity of the shape model \cite{Gross2005}. 
  Since labeling data for new identities can be labor intensive we would like to determine if AAM performance can be increased for an unseen identity using an identity-expression factorization extension to the traditional AAM formulation.  
  
  The traditional formulation on an AAM contains a PCA model of the joint point and appearance data. 
  This lends itself to a linear Gaussian interpretation which leads us to investigate whether key-point localization itself is improved using identity normalization. In this section we develop a novel identity-expression factorized AAM and investigate the utility of identity normalization for key point detection.
  \subsection{Inverse Compositional AAMs}
  The AAM fitting procedure we adapt is the inverse-compositional method as described by Matthews and Baker in \cite{matthews_active_2004}. We briefly review this method here. The modified objective in AAM fitting is to 
  minimize the error between a template image, $T$ and a set of warped images $I_j,j=1,2,...,J$, which generally belong to the same identity. The modified
  objective is to minimize the squared error between the set of image over all image locations $\x = (x, y)$, by estimating parameters $\p_j$ of a warp function $W$, or 
  \begin{equation} 
    \min_{\p_j} \sum_j \sum_\x [ T - I_j(W(\x; \p_j)) ]^2.
  \end{equation}
  We change notation here to make the following more clear.
  The warp function, $W$, can be interpreted as a map which determines a new location for any point $\x$, \ie, $\x' = W(\x, \p_j)$.
  For AAM models, this is typically a piecewise affine transform based on the triangulation of a shape $\s$, which are represented by $(x,y)$ point locations. 

  We shall quickly review how this optimization is performed with the classical AAM inverse compositional mapping approach, then replace the standard approach with our identity and expression factorization method and see the impact on the optimization procedure.
  In the standard AAM, $\p_j$ is the vector of weights for a set of eigen-shapes. The shape $\s$ is parameterized by a PCA model such that each $\s_j$ is explained by the mean shape $\s_0$ and a 
  weighted combination of eigen-vectors $\bf P_s$ with weight parameters $\p_j$, such that $\s_j = \s_0 + \mathbf{P}_s\p_j$. 

  The inverse compositional approach is to optimize this objective by iteratively building up a series of warps to recover an optimal warp $W(\x, \p_j)$. 
  Somewhat confusingly, both the template, $T$ and the image $I$ are warped. $T$ is always warped from its initial shape $\s_0$. $I$, however, is warped
  to the current estimate $I(W(\x, \p_j))$. Once the optimal warp at the iteration is determined, $\Delta \p_j$, the current warp parameters $\p_j$ and
  $\Delta \p_j$ are ``composed'' in order to update the current warp parameters. When $\Delta \p_j$ is close to zeros, or if the warp is not changing much,
  the algorithm has converged. The current warp is formed by inverting the parameters
  of $\Delta \p_j$ (by negating) and applying the warp defined by these updated parameters to a warp composed
  of all previous warps. The modified objective at each iteration is then
  \begin{equation} 
    \min_{\p_j} \sum_j \sum_\x [ T(W(\x; \Delta \p_j)) - I_j(W(\x; \p_j))]^2.
  \end{equation}
  Using a first order Taylor expansion, we have
  \begin{equation} 
    \min_{\p_j} \sum_j \sum_x [ T(W(\x; 0)) + \nabla T \frac{\partial W}{\partial \p} \Delta \p_j - I_j(W(\x; \p_j))]^2.
  \end{equation}
  So that $\Delta \p_j$ can be given as
  \begin{equation}
    \Delta \p_j = \hessian^{-1} \sum_\x \bigg[\nabla T \frac{\partial W}{\partial \p} \bigg][I_j(W(\x; \p_j)) - T(\x)], \label{update}
  \end{equation}
  for an individual $\p_j$, where $\hessian$, the Hessian, can be written as
  \begin{equation}
    \hessian = \sum_\x \bigg[\nabla T \frac{\partial W}{\partial \p} \bigg]^T\bigg[\nabla T \frac{\partial W}{\partial \p} \bigg].
  \end{equation}
  The algorithm consists of pre-computing $\nabla T$, $\frac{\partial W}{\partial \ve_j}$, $\hessian_v, \hessian_w$ and inverses at the mean shape for efficiency.
  These are used along with Eq. (\ref{update}) to compute the change at $(\p_j = 0)$ required to minimize the error
  between the template and the current warped target. To adapt this algorithm for our purposes requires few changes.
 
\subsection{A Novel Identity-Expression AAM (IE-AAM) Formulation}

  In the AAM inverse compositional method, one must determine $W(\x, \p^{k}) = W(\x, \p^{k-1}) \circ W(\x; \Delta \p)^{-1} \approxeq W(\x, \p^{k-1}) \circ W(\x; -\Delta \p)$ at each iteration $k$.
  To compute the parameters of  $W(\x, \p^{k})$ one then computes the corresponding changes to the current mesh vertex locations $\Delta \s = (\Delta x_1, \Delta y_1, \ldots, \Delta x_v, \Delta y_v)^T$,
  which are computed by composing the current warp.
  
  The new parameters at time $k$ are then given by $\p^k_j = \mathbf{P}_s^{-1}(\hat{\s}_j)$, where $\hat{\s}_j =  \s^{k-1}_j +  \Delta \s_j - \s_0$.
  Now, in the case of our identity and expression decomposition model, we replace the PCA model with a factorized model with the form $\s_j = \s_0 + \F\w + \G\ve_{j}$.
  The parameters are the vectors $\w$ and $\ve_j$ belonging to the identity and the $j^\text{th}$ expression of that identity. Note that compared to the notation used in Section \ref{sec:learning}, we have omitted the index $i$ since we are dealing with the same identity.
  In our identity and appearance decomposition approach, we now have an objective of the form, 
  \begin{equation} 
    \min_{\w,\ve_j} \sum_j \sum_\x [ T(\W(\x; \Delta \w, \Delta \ve_j)) - I_j(\W(\x; \w, \ve_j))]^2.
  \end{equation}
  Defining $\z_j = [\w^T,\ve_j^T]^T$, Taylor series expansion yields
  \begin{equation} 
    \min_{\w, \ve_j} \sum_j \sum_\x [ T(\W(\x; 0, 0)) + \nabla T \frac{\partial \W}{\partial \z_j} (\Delta \z_j) - I_j(\W(\x; \z_j))]^2,
  \end{equation}
  where $\Delta \z_j = [\Delta \w^T, \Delta \ve_j^T]^T$. As identity and expression are assumed to be independent, this can be computed separately,
  \begin{align*}
    \Delta \ve_j &= \hessian_{\ve}^{-1} \sum_\x \bigg[ \nabla T \frac{\partial \W}{\partial \ve} \bigg][I_j(\W(\x; \w, \ve_j)) - T(\x)], \\
    \Delta \w &= \hessian_\w^{-1} \sum_\x \bigg[\nabla T \frac{\partial \W}{\partial \w} \bigg] [\frac{1}{J} \sum_j (I_j(\W(\x; \w, \ve_j)) - T(\x))]. 
  \end{align*}
  %
  %
  We then compute the warp composition as in \cite{matthews_active_2004},
  \begin{align*} 
    W(\x, \w^k, \ve_{j}^k) &= W(\x, \w^{k-1}, \ve_j^{k-1}) \circ W(\x; -\Delta \w, -\Delta \ve_j).
  \end{align*}
  However, computing the parameters $\w^k$ and $\ve_j^k$ from $\Delta \s_j$ is no longer a simple matrix multiplication, as was the case when we updated $\p_j$ above,
  because the loading matrices of the identity expression model $\F$ and $\G$ are not orthonormal and we use the sequence of images. Therefore, we use the expectation of Eq. (\ref{posterior}).
  That is, after computing $\Delta s_j$ for each warp, let
  \begin{align}
    \dvect &\triangleq (\w^T, \ve^T_1, \ve^T_2, ..., \ve^T_J)^T, \\
    \hat{\s} &\triangleq ((\s^{k-1}_1 + \Delta \s_1 - \s_0)^T, ..., (\s^{k-1}_J + \Delta \s_J - \s_0)^T)^T.
  \end{align}
  Then, with $\bm{\Psi}$ constructed as a diagonal matrix with $\bm{\Sigma}$ along the diagonal repeated $J$ times, $\A_i$ as defined in Eq. (\ref{A}), and ${\bf \Phi}$ constructed as a diagonal matrix, by merging the  diagonal of $\lambda \I$ with $J$ blocks of diagonal of $\rho \I$ we can get
  \begin{align}
    \label{eq:AAM_posterior}
    \E[\dvect_i] &= ({\bf \Phi}^{-1} + \A^T{\Psi}^{-1}\A)^{-1}\A^T{\Psi}^{-1}(\hat{\s}^k)
  \end{align}
  with ${\bf \Phi}$ and $\A$ being equal to 
  \begin{align}
  \label{eq:AAM_phi}
    {\bf \Phi} & = diag\left(\lambda_{1},~\ldots,~\lambda_{K},~\rho_{1_{1}},~\ldots,~\rho_{1_{L}},~\ldots,~\rho_{J_{1}},~\ldots,~\rho_{J_{L}}\right),
    \end{align}
  \begin{equation}
  \label{AAM_A}
    \A = \left( \begin{array}{ccccc} 
      \F & \G &  0 & ... &  0 \\
      \F & 0 &  \G & ... &  0  \\
      \ldots & \ldots & \ldots & \ldots & \ldots \\
      \F & 0 &  0 & ... &  \G  \\    
    \end{array} \right)
  \end{equation}
  %
  %
  %
  from which we can then compute the current warp $W(\x, \w^k, \ve_{j}^k)$. 

  Texture variation is handled by alternating between training the model for texture variation and estimating the texture independently. Alternating between the two parameter
  optimization yielded the best results, although more efficient methods can be applied.
  
  \subsection{Keypoint Localization Experiments with CK+ and IE-AAMs}\label{sec:ck_pts}
  
  Since the CK+ dataset is also supplied with facial landmark annotations, we experimented using the Active Appearance Model extension described in previous sections. We used the same training set as
  described in the emotion recognition sections, training the factorized model on all but a single identity, and testing on those belonging to a single identity. 
  For all our experiments, we use leave-one-subject-out cross-validation of a dataset of 2588 images and point locations. 
  On average, this resulted in 2566 training images per experiment. The test-set were comprised of all remaining images from the CK+ dataset.

  We evaluate our approach by adapting an existing software package, the ICAAM software package \cite{Web:ICAAM} implemented in Matlab, replacing the PCA point distribution model with our factorized model,
  and comparing the results against the original software. Although more complex AAM software is available (for example the ICAAM package, which is not multi-resolution), our experiments are designed
  to illustrate both the effectiveness of the model and the simplicity of adapting existing approaches. 

  In order to achieve good results using this package, however, we used some pre-processing which improved ICAAM results:
  we first applied the OpenCV Viola-Jones frontal face-detector to remove large-scale pose variation and then aligned the images using a Procrustes analysis. The variation after Procustes alignment
  was modeled using PCA with 3 components and added to the shape projection matrix in ICAAM in order to model pose. In our case, 
  we added the pose parameters to the non-identity component matrix $\G$ and then ortho-normalization was applied, which follows Lucey \etal among others \cite{Lucey2010}.
  ICAAM was set to use the default 98\% of shape variation, while our method used 200 components for identity and 50 for expression.
  Test images were initialized using the face-detector used for training.

  \begin{figure}[htb]
    \centering
    \subfloat[Cumulative error distribution.]{
      \includegraphics[width=60mm]{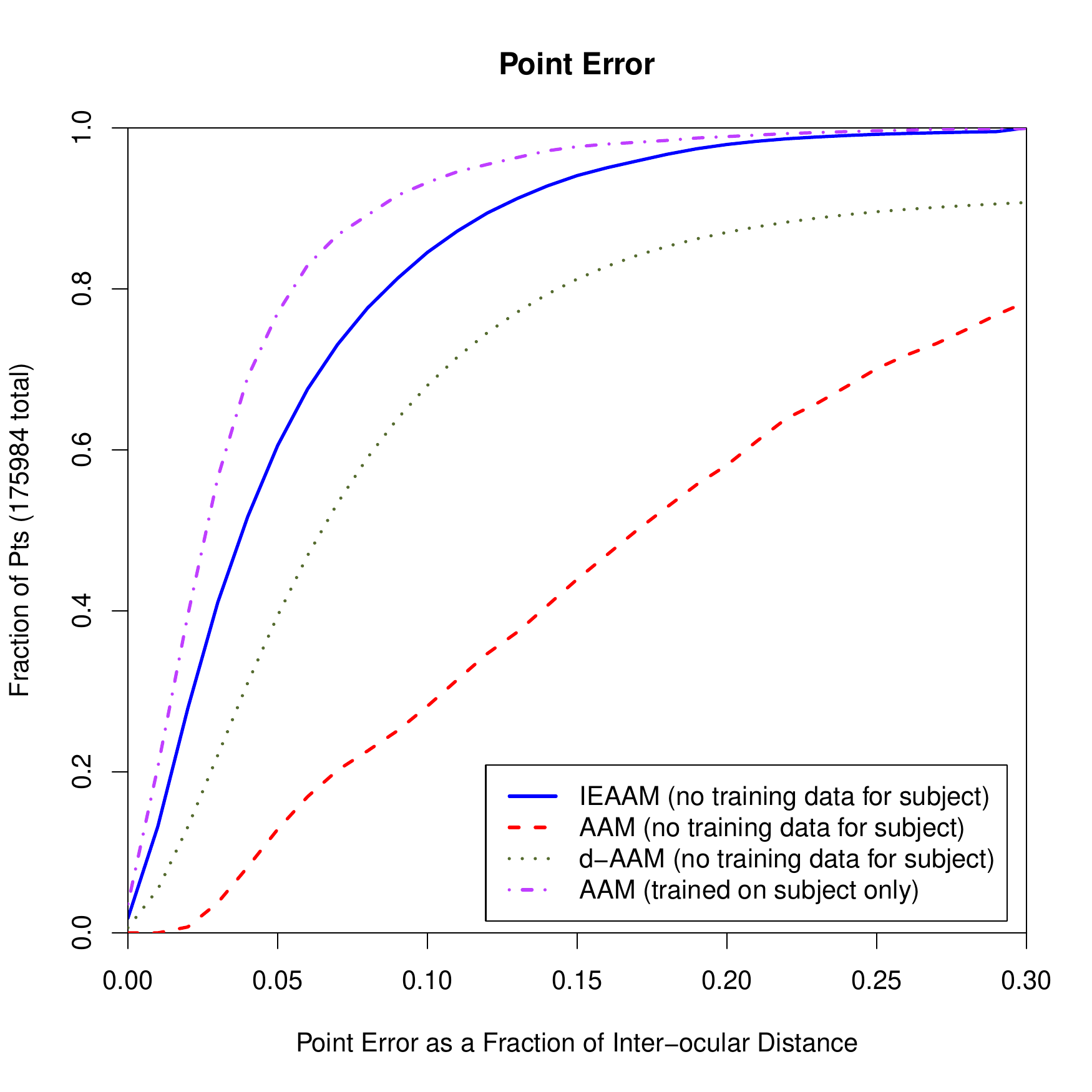}
      \label{fig:graphsb}
    }    
    \subfloat[Subject average error, as fraction of inter-ocular distance, cumulative distribution.]{
      \includegraphics[width=60mm]{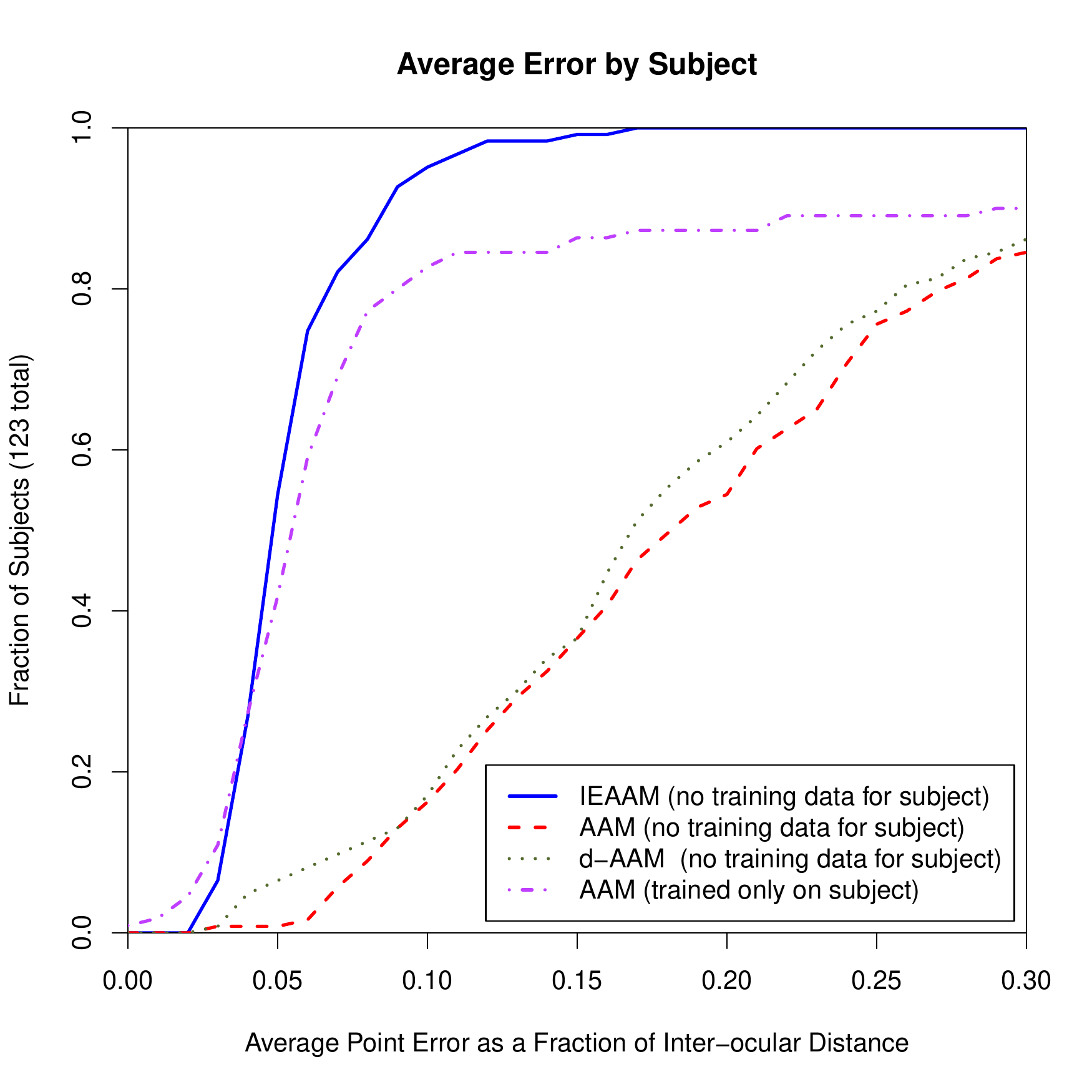}
      \label{fig:graphsa}
    }    
    \caption{Point-localization experiment evaluation, described in detail in text.}
    \label{fig:graphs}
  \end{figure}  
  The results are summarized in Figure \ref{fig:graphs}. 
  In \ref{fig:graphsb} we show the total cumulative distribution of error in all tested points for four methods. This analysis is widely used to evaluate keypoint techniques \cite{matthews_active_2004}. The Identity-Expression 
  AAM, (IEAAM), out-performs
  a discriminative AAM included in the Demolib distribution by Saragih \etal \cite{JasonSaragih2009} (d-AAM) and the ICAAM AAM on which our INM
  is based. In both cases, the default settings are used, with PCA dimensionality chosen
  to account for 95\% and 98\% of variation. These three methods are not trained on the left-out subject, but on all remaining identities.
  
  To provide another point of reference concerning what AAM is capable of achieving if training data is used for the subject of interest, we show the results of a subject-specific model using the AAM implementation provided in the DeMoLib distribution. This subject-specific AAM was trained on one sample of testing subject images described previously in Section \ref{sec:ck_au},
  and tested on all remaining subject images from the CK+ dataset. This corresponds
  to using the first and last images, along with a sampling from intermediary images in each sequence as training images. In CK+, 
  this corresponds to extreme expression poses. 
  This configuration simulates the labeling of a range of motion image sequence, which can be a good practical strategy for fitting an AAM with a small amount of labeling effort.
  The subject-specific AAM uses the project-out approach, \cite{matthews_active_2004} and the inverse-compositional method. As expected, this method performs 
  well, with less than 22.9\% of predicted points being greater than 5\% of the inter-ocular distance (calculated as the two farthest points on the eyes) from the ground truth.
  
  We also compute the average inter-ocular distance error for each subject. In \ref{fig:graphsa}, we show the cumulative distribution of average subject error, which evaluates
  how well methods perform within each subject trial. The subject-trained AAM suffers from convergence errors, \ie dramatic failures where all or many keypoints completely fall off the face, which cause the average error for each subject to 
  increase. In fact, the average error of all points for the subject-trained AAM is 23.3\% of inter-ocular distance, while the average error of our method is 5.94\%. Our method
  uses joint optimization on a sequence of frames adding stability to the point localization. Because of this, our method does not have these kinds of convergence errors, as shown in Figure \ref{fig:graphsa}, 

  Other methods, such as the one of Van der Maaten and Hendriks report 4.69 pixel error using mixture
  modeling on the CK+ data \cite{VanderMaaten2010}. They also report 0.23\% convergence error rate at that level of accuracy. More crucially, however, test subjects in their testing regimen were not strictly prohibited 
  from the training set. We are not aware of any AAM experiments using a large number of disparate subjects and testing on left-out subjects that report lower error. In any case, we are motivated
  more by the possibility of improving methods by using identity information, and believe that identity-normalization might easily be applied to any AAM-method. However, we do note that we see 0\% convergence error, which we believe is due to the inference over multiple images of the same identity.

  These results strongly suggest that identity is an important source of information and there are measurable benefits to modeling such information explicitly.
  The average error for the CK+ set using the ICAAM code with face detection initialization is 24.36, with a standard deviation of 11.10, 
  whereas our model yields an average pixel error of 7.15, with a standard deviation of 6.57. Both the
  bias and the variance is minimized simply by extricating the source of variation. The change to the algorithm, as discussed in previous sections, is relatively small.
  The result is significantly improved.
  
  \section{Identity-Expression Constrained Local Models (IE-CLMs)}
  \label{sec:IE-CLMs}
  
The term Constrained Local Models (CLMs) has evolved from the original work of Cristinacce and Cootes \cite{cristinacce2006feature}, which can be viewed as a particular instance of a CLM \cite{Saragih2010}. Nowadays the term CLM has come to refer to a number of methods which involve finding the landmarks of an image $\mathcal{I}$ through assigning a cost $\mathcal{Q}$ to candidate landmark positions $\{\bf {x}_1, ...,\bf {x}_n \} \in \bf {x}$ on the image and the parameters of the model $\bf {p}$.
The corresponding objective function can be written as:
\begin{align}
\label{eqn:CLM}
\mathcal{Q}(\bf {x}, \bf {p})=\sum_{i=1}^n \mathcal{D}_i(\bf {x}_i;\mathcal{I})+\mathcal{R}(\bf{x},\bf {p}),
\end{align}
where $\mathcal{D}_i$ encodes the image dependent suitability measure for the $i^{th}$ landmark being located at position $\bf{x}_i$ in the image. The $\mathcal{R}$ term can be interpreted as a regularization term that encodes preferences for
certain spatial configurations of landmark positions\footnote{We have reversed the order of terms compared to the notation in \cite{Saragih2010} and indicated an explicit dependence on $\bf{x}_i$ for the underlying objective and $\mathcal{R}$.}. This set-up leads to what some refer to as a deformable model fitting problem. Further, it is common to use a linear approximation for how the shape of non-rigid objects deform and a common variant of such a modeling technique is known as a point distribution model or PDM \cite{Cootes1995}. The PDM of \cite{Cootes1995} models non-rigid shape variations linearly and composes them with a global rigid transformation such that the 2D location of the PDM's $i^{th}$ landmark is given by:
\begin{align}
\label{eqn:PDM}
{\bf x}_i = s {\bf R}(\bm{\mu}_i + \bf {\Phi}_i \bf {q}) + \bf {t},
\end{align}
where the PDM parameters are defined by ${\bf p} = \{s,{\bf R},{\bf t},{\bf q}\}$. These parameters consist of a global scale $s$, rotation $\bf {R}$, and translation $\bf {t}$ (forming a similarity transformation), as well as a set of parameters encoded in vector $\bf {q}$ which capture the global non-rigid deformations though a sub-matrix $\bf {\Phi}_i$ of a larger matrix $\bf {\Phi}$ capturing variations using basis vectors. Using a probabilistic notation, the probability over the landmark position  ${\bf x}_i$ can be represented as a normal distribution with mean $s {\bf R}(\bm{\mu}_i + \bf {\Phi} _i\bf {q}) + \bf {t}$ and a covariance of ${\sigma}^2$, such that
\begin{align}
\label{eqn:prob_PDM_i}
p(\bf {x}_i) = \mathcal{N}( {\textit s} {\bf R}(\bm{\mu}_i + \bf {\Phi}_i \bf {q}) + \bf {t}, {\sigma}^2 ).
\end{align}
Consequently, the distribution over the whole set of landmarks ${\bf x}$ gets the form of a normal distribution which corresponds to 
\begin{align}
\label{eqn:prob_PDM}
p({\bf x}) = \mathcal{N}( {\textit s} {\bf R}(\bm{\mu} + \bf {\Phi} \bf {q}) + \bf {t}, {\sigma}^2 \bf{I} ),
\end{align}
where $\bm{\mu}$ and $\bf {\Phi}$ represent respectively a concatenation of the terms $\bm{\mu}_i$ and ${\bf \Phi}_i$ for $i \in \{1, \dots, n \}$. The Covariance ${\sigma}^2 \bf{I}$ is a diagonal matrix having the values in its diagonal being repeated $n$ times. 
Eq. (\ref{eqn:prob_PDM}) gives a principled probabilistic representation for the regularization term $\mathcal{R}$ in Eq. (\ref{eqn:CLM}).
 It is common to place a uniform prior on the parameters of the similarity transformation, while the model for non-rigid motion is typically represented using principal component analysis. 

\subsection{A Deeper Probabilistic View of PDMs}
When viewed through the lens of modern graphical modeling techniques and probabilistic principal component analysis, 
the classical formulation of PDMs can be re-written more formally as a probabilistic generative model over all landmarks $\bf{x}$, where
\begin{align}
p(\bf{x}) &=P(\bf{x}|\bf{z})P(\bf{z}|\bf{q})P(\bf{q}) \nonumber \\
&= \int_{\bf{q}} \int_{\bf{z}} 
\mathcal{N}(\bf{x}; \: {\textit s} \bf{R} \bf{z} + \bf{t}, \:\alpha\bf{I})
\mathcal{N}(\bf{z}; \:\bf{\Phi}\bf{q} + \bm{\mu},\: {\sigma}^2 \bf{I})
\mathcal{N}(\bf{q}; \: \bf{0}, \: \bf{I}) 
\: d\bf{q} \: d\bf{z}.
\end{align}
Parameter $\bf {q}$ here acts as the prior of the $p(\bf {x})$ distribution with mean zero and unit covariance matrix. Given $\bf {q}$, the conditional $P(\bf {z}|\bf {q})$ gets its mean by applying the non-rigid deformation $\bf {\Phi}$ and adding the $\bm{\mu}$ term. The mean of the conditional $P(\bf {z}|\bf {q})$ sets the intermediate value in the mean of Eq. (\ref{eqn:prob_PDM}) before applying the rigid transformations. Finally, the mean of the conditional $P(\bf {x}|\bf {z})$ is measured by applying affine transformations of rotation ${\bf R}$, scaling $s$, and translation {\bf t} on the variable $\bf {z}$. Having integrated out the variable $\bf {q}$, the distribution over all landmarks $\bf{x}$ can be simplified to
\begin{align}
p(\bf{x}) = \int_{\bf{z}} 
\mathcal{N}(\bf{x}; \: {\textit s} \bf{R} \bf{z} + \bf{t},\: \alpha\bf{I})
\mathcal{N}( \bf{z}; \: \bm{\mu}, \: \bf{\Phi}\bf{\Phi}^{T} + {\sigma}^2 \bf{I}) 
\: d\bf{z}.
\end{align}
Note that in this formulation the variable $\bf {z}$ represents the set of all landmarks in a transformed space where they are represented in a standardized coordinate space. That is different from their representation in $\bf{x}$ where each landmark is undergone an affine transformation and therefore can be observed in more volatile regions image. This difference guides us to a better representation in the $\bf {z}$ space to formulate the distribution of the key-points. Therefore, we transform the landmarks in each image from the $\bf{x}$ space to $\bf{z}$ space through a set of affine transformation, whose parameters are trained to be robust against translation, rotation, and scaling transformations. From this point forward, we discuss our model over the landmarks in the $\bf{z}$ space.

  \subsection{Probabilistic PCA based CLMs}
  \label{sec:PCA_CLM}
  Some of our previous work \cite{Hasan2013} has formulated a CLM in the following way; for a given image $\textit{I}$, 
%
we wish to combine the outputs of the local classifiers for the keypoints with a spatial model of global keypoint configurations. The local classifiers acts as discriminative predictors for the keypoints, while the spatial model provides a prior on how the keypoints are distributed.

\begin{figure}[htb]
    \centering
    \includegraphics[width=.5\textwidth]{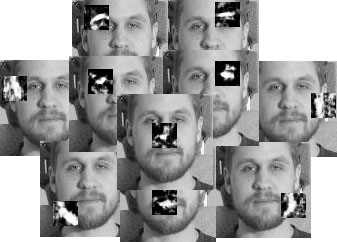}
    \caption{{Sample svm response maps generated by trained svms for an image. Each response map is generated by one local classifier trained for a particular keypoint.}} 
    \label{fig:svm_maps}
  \end{figure}

For each keypoint $i$ a local SVM classifier is trained. At test time, the local classifier for keypoint $i$, generates a response image map ${\bf d}^i$, which is a 2D array with a probability prediction for each pixel position in the image being the keypoint $i$. The set ${\bf D}=\{{\bf d}^1,{\bf d}^2 \cdots , {\bf d}^n\}$ represents the set of all generated response images for a given image, where $n$ is the number of keypoints. Figure \ref{fig:svm_maps} provides a visualization of the response image probability values. Note that the probabilities are scaled by a factor of 255 (8 bit gray-scale images).

Let ${\bf z}_i' \in {\bf d}^i$ indicate the coordinates of a gridpoint location on the 2D response image for keypoint $i$, then log of the score for the positive prediction of the local classifier at this location can be defined as $s_i({\bf z}_i')$. We use a probabilistic PCA over the keypoints to model the global keypoint configurations and use the log of its Gaussian distribution with a factorized covariance matrix as it energy term and couple it with the log score of the local classifier predictions in a spatial interaction energy function as follows:
\begin{align}
\label{eqn:joint_fitting}
& E(\bm{z}) =  
-\sum_{i=1}^{n}
\sum_{{\bf z}_i' \in {\bf d}^i}s_{i}({\bf z}_i')
\delta({\bf z}_{i}-{{\bf z}_i'}) \\
 & +
 \frac{1}{2}\left(\bf{z}-\bm{\mu}\right)^{T}
 (\bf{\Phi}\bf{\Phi}^{T} + {\sigma}^2\bf{I})^{-1}
 \left(\bf{z}-\bm{\mu}\right), \nonumber
\end{align}
where ${\bf z} = [{\bf z}_1, {\bf z}_2, \cdots {\bf z}_n ]^T $ gives the coordinates of $n$ candidate locations for the keypoints whose energy is measured. The enegy function $E$ is composed of two terms: 
The first term consists of the local response maps contribution where the first sum is over the keypoints and the second sum gets the log score $s_i({\bf z}_i')$ for each point ${\bf z}_i'$ in the set of all grid locations of the response image map ${\bf d}^i$. The terms $\delta$ is the dirac delta function whose output is one only if the grid location ${\bf z}_i'$ equals the queried location ${\bf z}_i$. 
The second term in Eq. (\ref{eqn:joint_fitting}) is log of the probabilistic PCA (PPCA) where $\bm{\mu}$ is simply the mean of the keypoints after RANSAC similarity registration. The terms $\bf{\Phi}$ and ${\sigma}^2$ in the covariance matrix of PPCA equal to:
\begin{align}
\label{eqn:PPCA_covariance}
\bf{\Phi} & = {\bf U}_p({\bf \Lambda}_p - {\sigma}^2\bf{I})^{1/2}{\bf R}, \\
{\sigma}^2 & = \frac{1}{n-p} \sum_{i=p+1}^{n} \lambda_i,
\end{align}
where ${\bf U}_p$ is a matrix of the $p$ principle eingenvectors of the keypoints in ${\bf z}$ space, ${\bf \Lambda}_p$ is a diagonal matrix whose diagonal is composed of eigenvalues $\lambda_1, \dots, \lambda_p$, knowing that eigenvalue $\lambda_j$ corresponds to the eigenvector in column $j$ of the matrix ${\bf U_p}$, and ${\bf R}$ is an arbitrary orthogonal matrix. The term ${\sigma}^2$ is the average of the remaining eigenvalues which explain the least significant direction of variation in data. Note that the first and second terms in Eq. (\ref{eqn:joint_fitting}) correspond respectively to the first and second terms of Eq. (\ref{eqn:CLM}).

To minimize $E$ we perform a search over the candidate keypoint locations $z$. This is done by iterating over the keypoints, where in each iteration all key-points are fixed except one. The optimum value for that keypoint is found through a comprehnsive search over the energy of the local response map grid locations. The iterations continue until all keypoints converge.



\subsection{Identity-Expression Factorized CLMs}
  Once the parameters of the identity expression model are learned using the EM procedure, as explained in Section (\ref{sec:learning}), this model is used in conjunction with the local response classifiers derived from the previously trained SVMs. Combining the energy of the local classifier with the energy term of the Identity Expression Factorized model, we minimize the energy of the following function:

\begin{align}
\label{eqn:energy_multiple_images}
& E(\bm{w}) =  
- \sum_{{\bf z} \in \bm{w}}
\sum_{i=1}^{n}
\sum_{{\bf z}_i' \in {\bf d}^i}s_{i}({\bf z}_i')
\delta({\bf z}_{i}-{{\bf z}_i'}) \\
& +
\left(\bm{w} - \bm{m} \right)^{T}
\left(\bm{\psi} + {\bf A} {\bf \Phi} {\bf A}^T\right)^{-1}
\left(\bm{w} - \bm{m} \right), \nonumber
\end{align}
  where $\bm{w}$ represents a set of candidate keypoints in all test images belonging to the same identity whose energy is measured. Note that as in the previous section $\bf z$ represents the set of all candidate keypoint locations on a single query image.
  Assuming there are a total of $\bm{J}$ images in $\bm{w}$ representing different expressions of a given identity, the term $\bm{m}$ is simply $\bm{J}$ times concatenation of $\bm{\mu}$, which brings the mean of the keypoints to a new space of dimensionality $\bm{J} \times n$.
  Compared to the first term in Eq. (\ref{eqn:joint_fitting}), there is an added summation which iterates over all images belonging to the same identity. The second term in Eq. (\ref{eqn:energy_multiple_images}), is equivalent to the log of the marginal distribution $P(\bm{w})$ of the identity expression factorized model, explained in Section~\ref{sec:basic_model}, which gets the probability of the joint set of points in all images of the same identity being the key-points. The term $\bm{\Psi}$ is constructed as a diagonal matrix by concatenating $J_i$ times the diagonal of $\Sigma$ and the matrix definitions of $\A$ and ${\bf \Phi}$ are given respectively in Eq. (\ref{AAM_A}) and Eq. (\ref{eq:AAM_phi}).
  
  The optimization procedure is done as follows; for each key-point $i \in n$ in each image $j \in \bm{J}$ a SVM response map ${\bf d}^i$ is generated. In each response map the position with the highest probability is chosen as the key-point location. The set of selected locations give the initial value of $\bm{w}$. Then, using Iterated Conditional Modes (ICM) \cite{besag1986statistical}, the energy of (\ref{eqn:energy_multiple_images}) is minimized by updating the position of each keypoint to it's minimum energy position while having all other keypoints fixed, iterating over all keypoints multiple times until no further updates are possible.
  
\subsection{Keypoint Localization Experiments with the MultiPIE Dataset and IE-CLMs}
\label{sec:IE-CLMs-expts}

\begin{figure}[ht]
\centering
\includegraphics[width=1.0\textwidth]{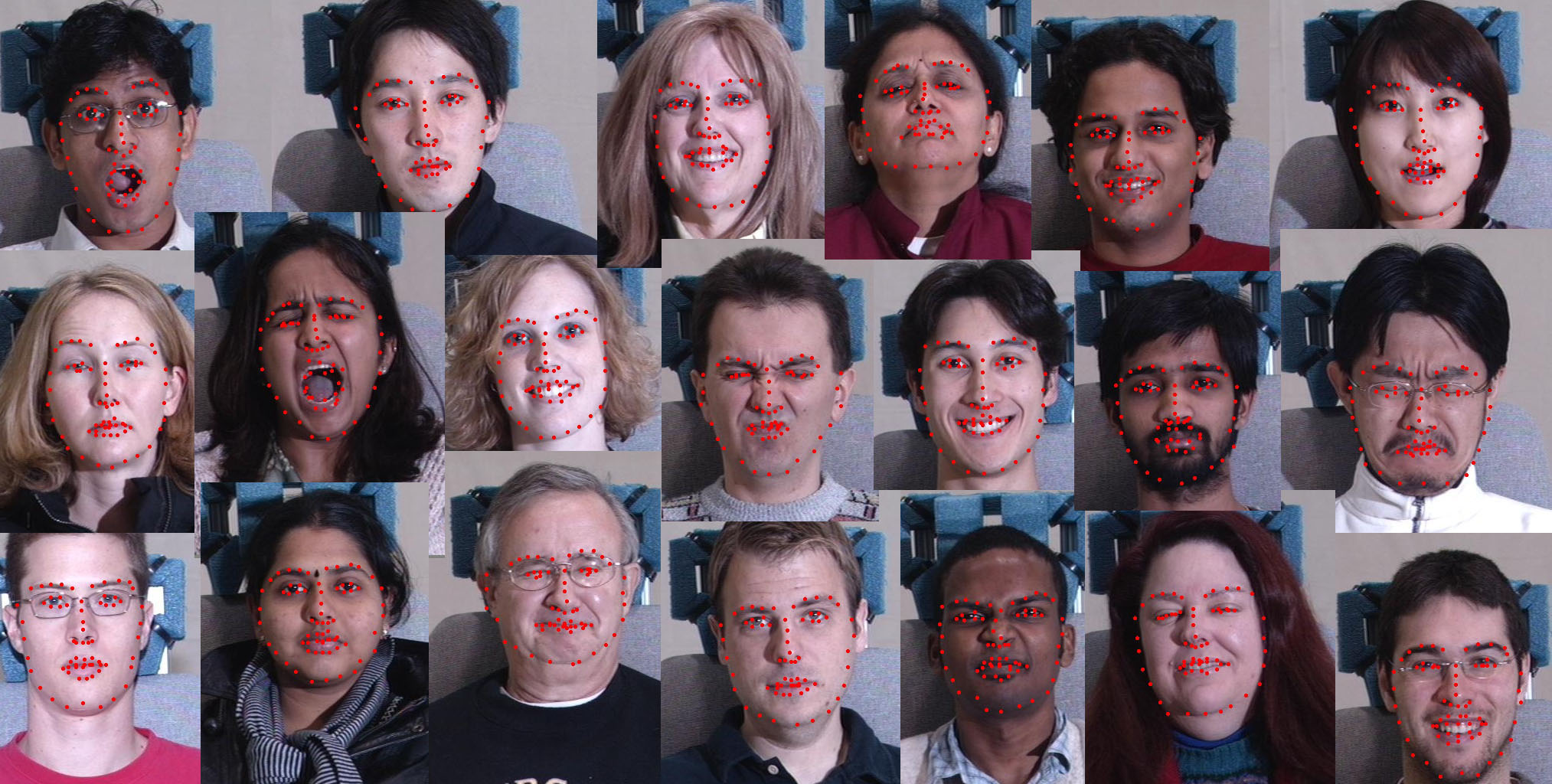}
\caption{{Sample of keypoint localization for multiple expressions of different identities by IE-CLM model.}} 
\label{fig:diff_id_exp}
\end{figure}  

\begin{figure}[ht]
\captionsetup[subfloat]{labelformat=empty}
\begin{center}
\subfloat[1st (left) and 2nd (right) directions for id.]{{
\includegraphics[width=0.22\textwidth, height=3.1cm]{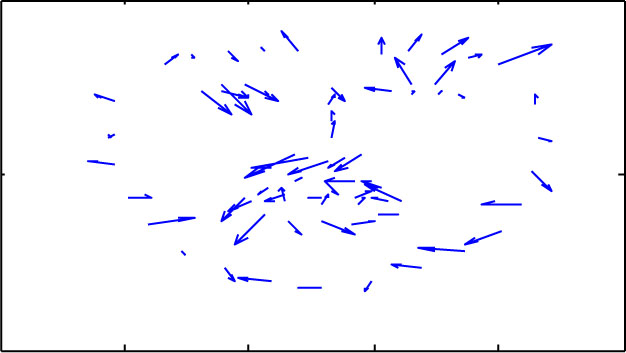}}
\subfloat{}{
\includegraphics[width=0.22\textwidth, height=3.1cm]{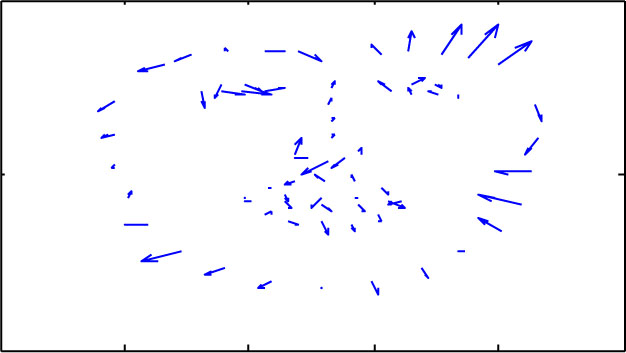}
}}
\hspace{1em}
\subfloat[1st (left) and 2nd (right) directions for exp.]{{
\includegraphics[width=0.22\textwidth, height=3.1cm]{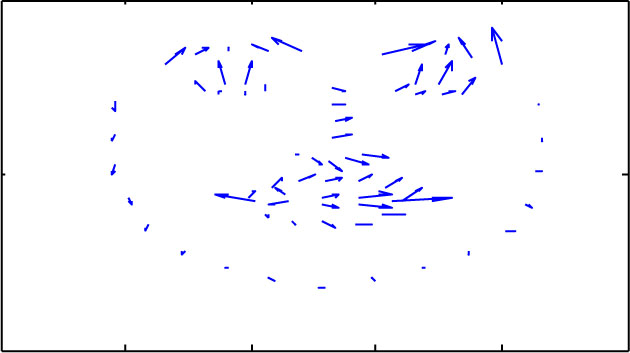}}
\subfloat{}{
\includegraphics[width=0.22\textwidth, height=3.1cm]{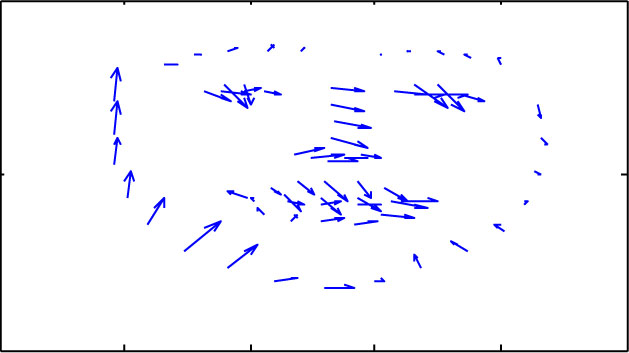}
}}
\caption{The main directions of variation across identities and expressions}
\label{fig:dir-variance}
\end{center}
\end{figure}

The CMU MultiPie face dataset \cite{MultiPie} captures a subset of expressions, such as surprise, smile, neutral, squint, disgust, and scream, of 337 identities. The images are taken in four sessions and in each session a subset of expressions is captured. Therefore, not all expressions are registered per identity. The availability of multiple expressions per identity provides the right dataset to evaluate our identity expression model. We perform the same experiment as in  \cite{zhu2012face} and compare our model with other models on the frontal face images with 68 keypoints. Following the same split of the dataset as in \cite{zhu2012face}, the first 300 images are used for training and the next 300 images are used for testing. 
Figure \ref{fig:dir-variance} shows the main directions of variation for the identity and expression features captured by our model on MulitPie training set. The directions of variation for expression captures how face changes in between different emotions. On the other hand, the directions of variation for identity mainly deals with different face sizes as well as some 2D in-plane rotation which is due to the fact that the faces are not completely frontal in some of the expressions of an identity. 

Using the training set of MultiPie, we train a binary SVM classifier on HOG feature representation for each keypoint. Indeed, for training the SVM classifier of each keypoint, positive patches are generated from the training set centered at the keypoint and negative patches are generated within a bounding box around the keypoint.
%
We compare our model on MultiPie test-set with the independent model of Zhu and Ramanan \cite{zhu2012face} and other models reported in that paper as well as the more recent model of Hasan \textit{et al.} \cite{Hasan2013}. The error is the average difference in euclidean space between the true pixel locations and the predicted ones normalized by the face size, which is the average of height and width.
The results are reported in Table \ref{tab:localization_error}. Our model catches up with Zhu's model at 0.06 of face fraction size, though at face fraction size of 0.05 the difference is marginal at 0.016. The values of the CLM model of Saragih \textit{et al.} \cite{Saragih2010} are reported from \cite{zhu2012face} where the latter source evaluates the previously trained CLM model on MultiPie test-set. However, it is not clear on which data the CLM model was trained on and which feature representation was used for training them. 
In order to provide a CLM model trained on precisely the same training set and using precisely the same features and local classifiers, we use the PPCA-CLMs described in Section~\ref{sec:PCA_CLM}, which are trained on HOG features. This allows us to establish a more highly controlled baseline for the PCA vs IE-CLM comparison. The performance of our PPCA-CLM is different from the PCA-CLM of Saragih \textit{et al.}. These differences could be due to the difference in the training set used, or due to the difference in the feature representation. Other variations could arise from our slightly different underlying CLM formulation or our use of ICM as the inference procedure. We see that for small face fraction size accuracy the difference is significant; however, at the 0.06 face fraction threshold, the performance of the two PCA based CLM methods are comparable.

\begin{figure}[htb]
\centering
\subfloat{
\includegraphics[height=29mm]{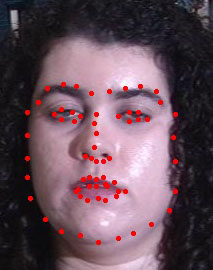}
}
\subfloat{
\includegraphics[height=29mm]{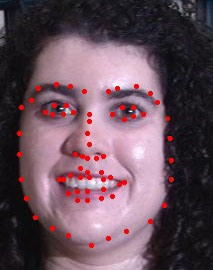}
}
\subfloat{
\includegraphics[height=29mm]{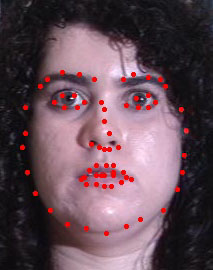}
}
\subfloat{
\includegraphics[height=29mm]{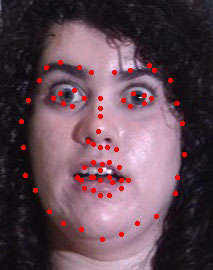}
}
\subfloat{
\includegraphics[height=29mm]{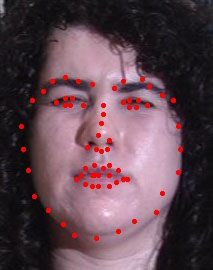}
}\\
\subfloat{
\includegraphics[height=29mm]{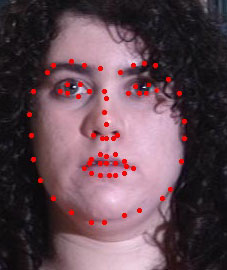}
}
\subfloat{
\includegraphics[height=29mm]{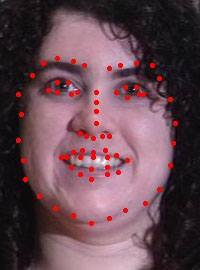}
}
\subfloat{
\includegraphics[height=29mm]{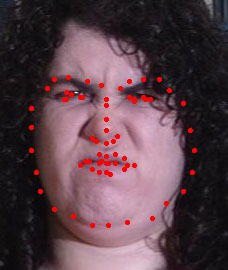}
}
\subfloat{
\includegraphics[height=29mm]{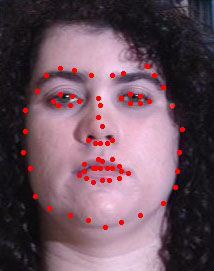}
}
\subfloat{
\includegraphics[height=29mm]{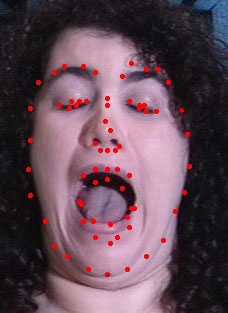}
}
\caption{Labeled images by Identity Expression Factorized CLM model for different expressions of identity sample 1.}
\label{fig:Expr}
\end{figure}

\begin{figure}[htb]
    \centering
    \subfloat{
      \includegraphics[height=29mm]{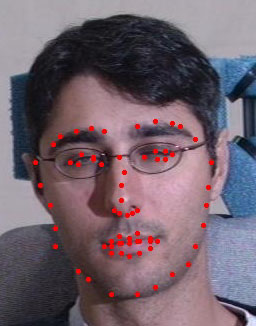}
    }
    \subfloat{
      \includegraphics[height=29mm]{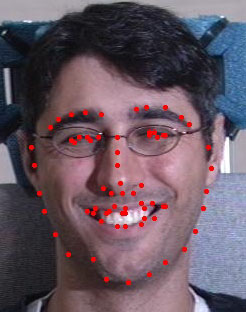}
    }
    \subfloat{
      \includegraphics[height=29mm]{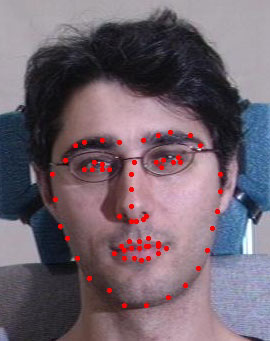}
    }
    \subfloat{
      \includegraphics[height=29mm]{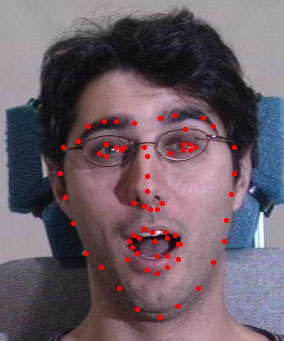}
    }
    \subfloat{
      \includegraphics[height=29mm]{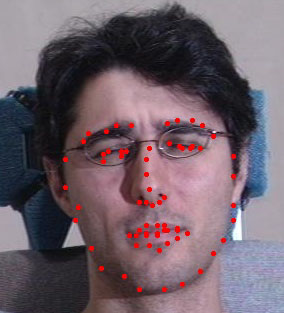}
    }\\
    \subfloat{
      \includegraphics[height=29mm]{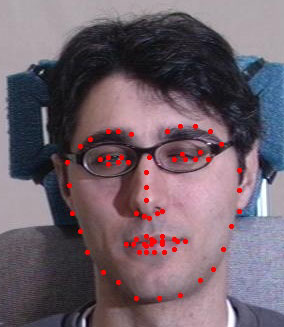}
    }
    \subfloat{
      \includegraphics[height=29mm]{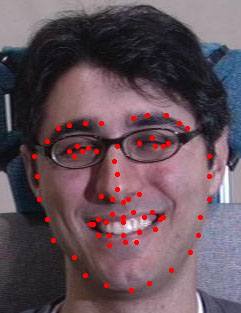}
    }
    \subfloat{
      \includegraphics[height=29mm]{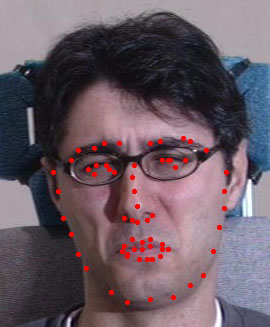}
    }
    \subfloat{
      \includegraphics[height=29mm]{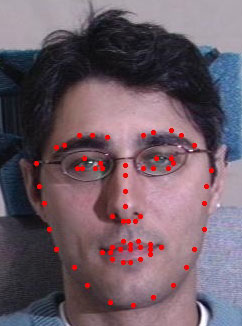}
    }
    \subfloat{
      \includegraphics[height=29mm]{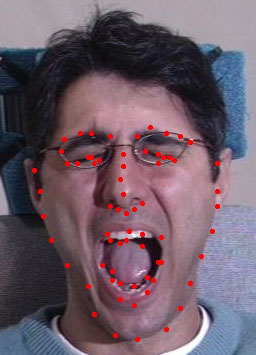}
    }
    \caption{Labeled images by Identity Expression Factorized CLM model for different expressions of identity sample 2.}
     \label{fig:expr}
  \end{figure}

Turning to the IE-CLM we see that it lags behind Saragih's CLM only at the very small face fraction accuracy level. We speculate that the lower performance of IE-CLM in this range is due to the difference in the training data for the local classifiers. The IE-CLM model catches up rapidly to both Saragih's CLM and the Multi-AAM \cite{multiAAM} models from the 0.04 face fraction level forward, and yields predictions for 100\% of the points with an error of less than 0.06 of the face size. This level of performance is on par with the state of the art methods of \cite{zhu2012face} and \cite{Hasan2013}.
Figures \ref{fig:Expr} and \ref{fig:expr} illustrate our model's performance on different expressions of two sample identities in the test-set. Our model exploits the fact that all these images have the same identity and thereupon, the same identity representation is used when the model searches for the optimal keypoints. Figure \ref{fig:diff_id_exp} shows some sample keypoint localization by our model on different identities in the test set, which includes data of different expression, ethnicity, age, gender, illumination, and face masks such as trimming style and eye-glasses usage. 

In terms of running time, the PPCA-CLM model takes 32 seconds for the local response image generation and 6 seconds for the optimization procedure. However, we have use none of the obvious acceleration techniques that could be applied to speed the local response map generation procedure, such as GPU acceleration or optimizing operations that are convolutational in nature. The IE-CLM model's time changes linearly based on the number of images of the identity being evaluated. If intended for an interactive or real-time procedure where the images are to be processed sequentially, the timing could be made much more comparable to a PCA-CLM for which various real-time implementations are available. To accelerate the IE-CLM computations one could adapt the joint inference procedure over all images in a batch so as to \emph{incrementally} use estimated factors from images in previous time steps. This could be formulated as a form of incremental inference or probability propagation, an approach which is fairly well understood in the literature. 
%

%
\begin{table}
  \centering
  \caption{Percentage of faces with an average localization error less than the given fraction of face size on MultiPie dataset.}
  \vspace{5pt}
  \begin{tabular}{l c c c c}
Fraction of face size & 0.03  & 0.04 & 0.05  & 0.06  \\
    \hline
    Zhu and Ramanan (independent model) \cite{zhu2012face}  
    & 0.86    & 0.97    & 1.00    & 1.00  \\
    Hasan \textit{et al.}   \cite{Hasan2013} 
    & 0.78	  & 0.94	& 1.00	  & 1.00  \\
    Oxford	\cite{oxford} 
    & 0.28	  & 0.77	& 0.94	  & 0.98  \\
    Star Model  \cite{star}
    & 0.29	  & 0.80	& 0.92	  & 0.92  \\
    Multi-AAM	\cite{multiAAM}
    & 0.64	  & 0.87	& 0.91	  & 0.92  \\
    CLM of Saragih \textit{et al.} 	\cite{Saragih2010}  
    & 0.68  & 0.85	& 0.90	  & 0.93  \\
    Face.com \cite{face}	
    & 0.31	  & 0.61	& 0.79	  & 0.87  \\
    \hline
    Our PPCA-CLM	& 0.13	  & 0.54	& 0.79	  & 0.92  \\
    Our IE-CLM	  & 0.52    & 0.89    & 0.98    & 1.00    \\
    \hline
  \end{tabular}
  \label{tab:localization_error}
\end{table}

  \section{Conclusions and Future Work}

  We have shown that identity-expression disentanglement and various forms of identity normalization can be used to improve the performance of supervised learning approaches to performance-driven animation and expression recognition tasks. Detailed facial expression analysis techniques are often limited by the need for subject-specific models, which limits the quantity of labeled data that can be brought to bear.
  In our work here, we have detailed how approaches that work well for a single subject can be adapted and their performance improved for unseen identities. Our approach is based on disentangling identity and expression with data that was been weakly labeled according to identity. The increase in performance across many tasks suggests that using simpler tools for both performance-driven animation and automatic expression recognition may be sufficient if identity labels can be used as an additional source of information. 

  In future work it might be interesting to extend this type of approach to be able to deal with more complex data, including data exhibiting greater pose variation. For example an explicit 3D shape model such as the formulation proposed for CLMs in the work of Jeni et al. \cite{Jeni} could be combined with the identity-expression factorization approach used here. 
  
  We have shown how the identity-expression factorization approach can be integrated into a CLM for keypoint localization. We believe there are a number of possible directions that could lead to further improvements. 
  Firstly, we believe that replacing the binary SVM classifier with a multi-class SVM might yield increased performance with minimal changes required to the underlying model formulation. However, once formulated in this way, it would also be more straight-forward to train the model to make structured predictions using a fully discriminative training procedure, for example, along the lines of \cite{zhu2012face}.
  Another direction is to replace the Iterated Conditional Modes (ICM) energy minimization step with a temperature based annealing procedure. 
  The ICM algorithm is known to be quite fast at finding local minima, but it may be the case that better solutions could be found with a stochastic annealing procedure.

The use of multi-scale intensity response information in conjunction with alternative optimization techniques could be a promising direction for future research. It is also likely possible to speed up response image generation and energy minimization when multiple images are processed per identity through various means.

  Finally, the local SVM classifiers and hand engineered features that we have used here could be replaced with a convolutional neural network (CNN), such as the architecture proposed in \cite{honari2016CVPR}. A CNN approach was not feasible in the context of our primary goal of animation control at the time this work was performed since CNNs typically require a considerable amount of training data to avoid overfitting and our face camera dataset is quite small compared to the data sets that have been used where CNNs have yielded state of the art results. However, for other application settings and datasets research combining identity-expression disentanglement techniques with CNNs may be a promising direction of future research. 

  \section*{Acknowledgements}
  We thank Ubisoft for both financial support and for providing the helmet camera video data used for our high quality animation control experiments. We also thank the Natural Sciences and Engineering Research Council of Canada (NSERC) for financial support under the CRD and Discovery grant programs and the Fonds de recherche du Québec - Nature et technologies (FRQNT) for a Doctoral research scholarships (B2) grant to SH. We thank Yoshua Bengio for his comments and suggestions regarding this work.

  \bibliographystyle{ieee}
  \bibliography{imavis}
  
  
  
  
  
  

\end{document}